\documentclass[5p,times,twocolumn]{elsarticle}
\usepackage{setspace}
\usepackage[switch]{lineno}
\usepackage[normalem]{ulem}

\usepackage{comment}
\usepackage{amsmath}
\usepackage{graphicx}
\usepackage{hyperref}
\usepackage{float}
\usepackage{subcaption}
\usepackage{multirow}
\usepackage{afterpage}
\usepackage{etoolbox}
\usepackage[english]{babel}
\usepackage{natbib}
\usepackage{hhline,colortbl}
\usepackage{tablefootnote}

\definecolor{RevisionColor}{rgb}{0.0, 0.0, 0.0} 
\newcommand{\revision}[1]{\color{RevisionColor}{#1}\normalcolor}

\definecolor{Revision2Color}{rgb}{0.0, 0.0, 0.0}
\newcommand{\revisionTwo}[1]{\color{Revision2Color}{#1}\normalcolor}

\definecolor{todoColor}{rgb}{0.0,0.0,1.0}

\definecolor{todoColorOthers}{rgb}{1.0,0.0,0.0}

\definecolor{noteToSelfColor}{rgb}{0.6,0.6,0.6}
\newcommand{\noteToSelf}[1]{\color{noteToSelfColor}[{#1}]\normalcolor}

\definecolor{todoDoneColor}{rgb}{1.0,0.5,0.8} 
\newcommand{\todoDone}[1]{\color{todoDoneColor}[Check: {#1}]\normalcolor}

\definecolor{unresolvedColor}{rgb}{0.6,0.0,0.0}

\definecolor{questionColor}{rgb}{1.0,1.0,0.0}

\definecolor{answerColor}{rgb}{0.0,1.0,0.0}
\newcommand{\answer}[1]{%
  \colorbox{answerColor}{%
    \begin{minipage}[t]{\linewidth -2\fboxsep}%
        {ANSWER: #1}%
     \end{minipage}%
  }%
}

\newcommand{\fat}[1]{\mathbf{#1}} 
\newcommand{\bldgr}[1]{\boldsymbol{#1}} 
\newcommand{\transp}{^T} 

\newcommand{\argmax}{\operatornamewithlimits{argmax}}
\newcommand{\argmin}{\operatornamewithlimits{argmin}}

\setcitestyle{authoryear,round,semicolon} 

\let\today\relax
\makeatletter
\def\ps@pprintTitle{%
    \let\@oddhead\@empty
    \let\@evenhead\@empty
    \def\@oddfoot{\footnotesize\itshape
         {Submitted preprint to ???} \hfill\today}%
    \let\@evenfoot\@oddfoot
    }
\renewcommand{\fnum@figure}{Fig. \thefigure}
\addto\captionsenglish{\renewcommand{\appendixname}{Appendix}}
\makeatother

\begin{document}

\hyphenation{SAMSEG}

\begin{frontmatter}

\title{
An Open-Source Tool
for
Longitudinal 
Whole-Brain and White Matter Lesion Segmentation
}

\author[add1]{Stefano Cerri\corref{cor1}}
    \ead{scerri@mgh.harvard.edu}
\author[add1,add3]{Douglas N. Greve}
\author[add1]{Andrew Hoopes}
\author[add2]{Henrik Lundell}
\author[add2,add4,add5]{Hartwig R. Siebner}
\author[add6]{Mark M\"{u}hlau}
\author[add1,add7]{Koen Van Leemput}

\address[add1]{Athinoula A. Martinos Center for Biomedical Imaging, Massachusetts General Hospital, Harvard Medical School, USA}
\address[add2]{Danish Research Centre for Magnetic Resonance, Copenhagen University Hospital Amager and Hvidovre, Copenhagen, Denmark}
\address[add3]{Department of Radiology, Harvard Medical School, USA}
\address[add4]{Department of Neurology, Copenhagen University Hospital Bispebjerg and Frederiksberg, Copenhagen, Denmark }
\address[add5]{Institute for Clinical Medicine, Faculty of Medical and Health Sciences, University of Copenhagen, Denmark}
\address[add6]{Department of Neurology and TUM-Neuroimaging Center, School of Medicine, Technical University of Munich, Germany}
\address[add7]{Department of Health Technology, Technical University of Denmark, Denmark}

\cortext[cor1]{Corresponding author}

\begin{abstract}

In this paper we 
describe and validate
a longitudinal 
method for whole-brain
segmentation of longitudinal MRI scans. 
It 
builds upon
an existing
whole-brain segmentation 
method
that can handle multi-contrast data and 
robustly analyze images with white matter lesions.
%
This method
is here extended with subject-specific latent variables that encourage temporal consistency between 
its segmentation results,
enabling 
it
to 
better
track 
subtle 
morphological changes in 
dozens of neuroanatomical structures
and white matter lesions.
%
%
%
%
We validate the proposed method on multiple datasets 
of control subjects and patients suffering from Alzheimer's disease and multiple sclerosis,
and compare its results against those obtained with its original cross-sectional 
formulation
and two benchmark longitudinal methods.
The results indicate that 
the
method 
attains
a higher test-retest reliability, while being more sensitive to 
longitudinal disease effect differences between patient groups. 
An implementation 
is publicly available as part of the open-source neuroimaging package FreeSurfer.\\


\end{abstract}

\end{frontmatter}


\section{Introduction}
\label{sec:Intro}

Longitudinal imaging studies, in which subjects are scanned repeatedly over time, have several advantages over cross-sectional studies. Accordingly, longitudinal neuroimaging studies have provided valuable insights into temporal changes in healthy brain development~\citep{Giedd1999, Evans2006, Brain2012, Choe2013, Mills2021} and aging~\citep{Scahill2003, Tamnes2013}, as well as from neurodegenerative diseases such as Alzheimer’s disease (AD)~\citep{Fox1996, Laakso1998, Du2001, Halliday2017} or multiple sclerosis (MS)~\citep{Audoin2006, Fisher2008}. In most instances, a longitudinal study design increases statistical power compared to a cross-sectional design. Furthermore, only longitudinal studies allow for a reliable evaluation of interventions such as treatment effects. Most importantly, only longitudinal measures allow for monitoring the individual patient. In neuroimaging, preprocessing tools are commonly designed for cross-sectional data so that their use in longitudinal data may not fully exploit the advantages of the longitudinal study design with the risk of an overestimation of statistical power or the need of a higher number of subjects, respectively. 

\begin{figure*}[!ht]
    \centering
    \includegraphics[width=0.9\linewidth]{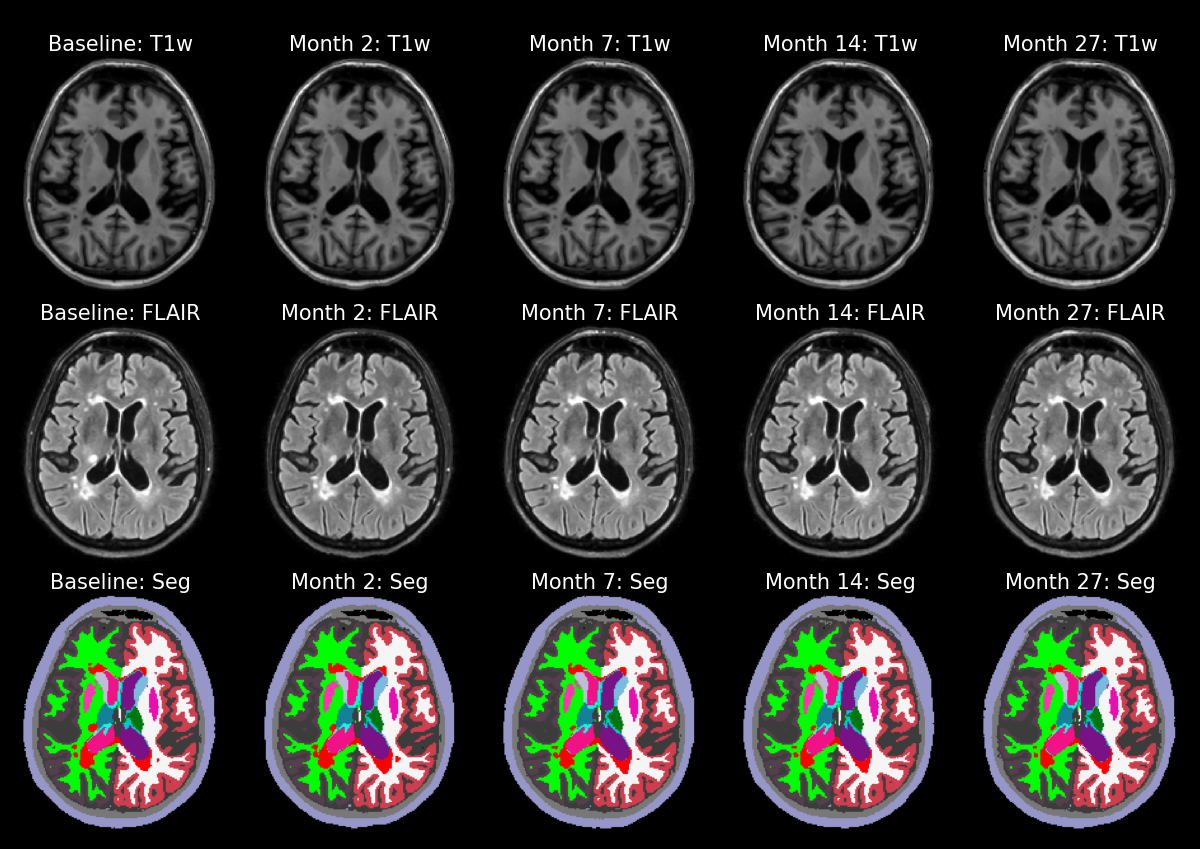}
    \caption{Whole-brain and white matter lesion segmentations (labeled in red) produced by the proposed method from T1w and FLAIR longitudinal scans of an MS patient.
    \revision{T1w=T1-weighted, FLAIR=FLuid Attenuation Inversion Recovery, MS=Multiple Sclerosis.}
    }
    \label{fig:segExample}
\end{figure*}

Over the last few decades, many dedicated neuroimage analysis tools have been developed to handle longitudinal data. These methods aim to exploit the expected temporal consistency in longitudinal scans to
obtain more sensitive 
measures
of longitudinal changes than is possible by analyzing each time point 
separately.
One class of algorithms is designed to detect changes between two consecutive time points without explicitly \emph{segmenting} each scan. These methods work by subtracting the two images to highlight locations of change~\citep{Hajnal1995, Freeborough1997, Lemieux1998, Battaglini2014}, or, more generally, by tracking corresponding voxel locations over time using 
nonlinear registration, and analyzing the estimated spatial deformations~\citep{Thirion1999, Rey2002, Avants2007, Holland2011, Elliott2019}.
Another class of methods explicitly \emph{segments} each time point in longitudinal scans, enforcing temporal consistency either on the segmentations themselves\revision{~\citep{Metcalf1992, Solomon2004, Xue2005, Xue2006, Wolz2010, Dwyer2014,Wei2021}}
or on spatial probabilistic atlases that are used to compute them~\citep{Shi2010, Shi2010b, Prastawa2012, Aubert-Broche2013, Iglesias2016, Tustison2019}. In order to make the various time points comparable on a voxel-based level, these methods typically involve a temporal registration step, computed either prior to~\citep{Wolz2010, Aubert-Broche2013, Gao2014, Iglesias2016, Tustison2019, Schmidt2019, Wei2021} or simultaneously with~\citep{Xue2005, Xue2006, Shi2010, Shi2010b, Li2010, Wang2011, Prastawa2012, Wang2013} the segmentations.
%

%
To date, most methods for analyzing longitudinal scans are designed to compute only very specific outcome variables, such as change in overall brain size~\citep{Hajnal1995, Freeborough1997, Smith2001, Smith2002} 
or global white/gray matter volume~\citep{Xue2005, Xue2006, Shi2010, Shi2010b, Prastawa2012, Gao2014},
cortical thickness~\citep{Nakamura2011, Wang2011, Wang2013, Tustison2019}, white matter lesions~\citep{Gerig2000, Solomon2004, Elliott2013, Schmidt2019, Birenbaum2016, McKinley2020, Sepahvand2020, Denner2020} or individual brain structures such as the hippocampus~\citep{Wolz2010, Iglesias2016, Wei2021}.
%
To the best of our knowledge, the most comprehensive tool for longitudinal analysis of structural brain scans is currently the one distributed with FreeSurfer~\citep{Reuter2012, Fischl2012}.
This tool segments many neuroanatomical structures simultaneously 
(%
both 
volumetric ``whole-brain'' segmentations
and
parcellations of the cortical surface%
),
and 
can readily handle 
data with more than two time points.
However, 
it is 
specifically 
designed 
for
T1-weighted (T1w) scans only 
--
as such it is 
less
well suited 
for studying
populations
with
white matter lesions and other pathologies
that
are better visualized using other MRI contrasts (such as T2w or FLAIR).
%
%
Furthermore, 
a 
recent
study 
suggests
that, even in T1w images,
it may be less sensitive to 
longitudinal changes 
than
the method 
we describe here~\citep{Sederevivcius2021}.

The 
contribution
of this paper is twofold. 
First, 
we 
make
publicly available
a new
method for 
automatically segmenting dozens of neuroanatomical structures from
longitudinal scans, 
using a model-based approach that
can take multi-contrast data as input and that can also segment white matter lesions simultaneously.
%
%
The method is fully adaptive to different MRI contrasts and scanners, and does not put any constraints on the number or the timing of longitudinal follow-up scans. 
%
%

Second,
we conduct an extensive validation of the proposed tool
using over \revision{4,500} brain scans acquired with different scanners, field strengths and acquisition protocols, involving both controls and patients suffering from MS and AD.
We demonstrate experimentally that the method produces more reliable segmentations 
in scan-rescan settings 
than 
the 
longitudinal tool in FreeSurfer
and
than
a cross-sectional 
version of the method,
while also being 
more sensitive to differences in longitudinal changes between patient groups.

An example of longitudinal segmentations of an MS patient produced by the proposed method is shown in Fig.\ref{fig:segExample}.
A preliminary version of this work, with a 
limited validation, appeared earlier as a workshop paper~\citep{Cerri2020}.

\section{Existing cross-sectional method -- SAMSEG}
\label{sec:CrossModel}

\begin{figure*}[!ht]
    \centering
    \includegraphics[width=0.85\textwidth]{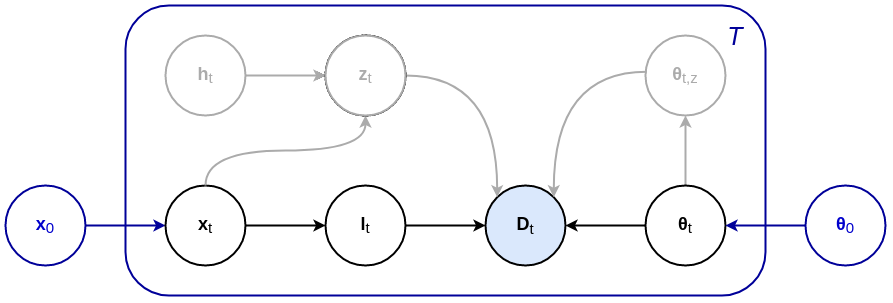}
    \caption{
    Graphical representation of the proposed longitudinal generative model.
    For each time point $t$, in black the cross-sectional model of \citep{Puonti2016}, in which image intensities $\fat{D}_t$ are generated from likelihood parameters $\bldgr{\theta}_t$ and segmentation labels $\fat{l}_t$ -- which in turn have been generated from an atlas-based segmentation prior with node positions $\fat{x}_t$.
    %
    In blue, the proposed additional subject-specific latent variables that encourage temporal consistency between longitudinal scans in the segmentation prior (through $\fat{x}_0$) and in the likelihood function (through $\bldgr{\theta}_0$).
    Also shown, in gray, is the cross-sectional lesion extension of \citep{Cerri2021}. For each time point $t$, $\fat{z}_t$ is a binary white matter lesion segmentation, $\fat{h}_t$ are latent variables encoding lesion shape information, and $\fat{\theta}_{t,z}$ are lesion intensity parameters constraining lesion appearance. 
    Shading indicates observed variables, while the plate indicates $T$ repetition of the included variables.
    }
    \label{fig:model}
\end{figure*}

We build upon a previously validated cross-sectional method for whole-brain segmentation called Sequence Adaptive Multimodal SEGmentation (SAMSEG)~\citep{Puonti2016}. 
%
SAMSEG segments 41 anatomical structures from brain MRI, and it is fully adaptive to different MRI contrasts and scanners. We here briefly describe the method as we extend it for longitudinal scans in the remainder of the paper.

Let $\fat{D} = ( \fat{d}_{1}, \dots , \fat{d}_{I} ) $ be the image intensities of a multi contrast scan with $I$ voxels, where $\fat{d}_{i} = ( d_i^1, \dots , d_i^N )\transp $ is the vector containing the log-transformed image intensities of voxel $i$ for all the available $N$ contrasts.
Furthermore, let $\fat{l} = (l_1, \dots , l_I)\transp$ be the corresponding segmentation labels, where $l_i \in \{1, \dots,  K\}$ denotes one of the $K$ possible anatomical structures assigned to voxel $i$. 
In order to compute segmentation labels $\fat{l}$ from image intensities $\fat{D}$, we use a generative model illustrated in black in Fig.~\ref{fig:model}.
It defines a 
forward model composed of two parts: a segmentation prior $p(\fat{l} | \fat{x})$, with parameters $\fat{x}$, that encodes spatial information of the labels $\fat{l}$, and a likelihood function $p(\fat{D} | \fat{l}, \bldgr{\theta} )$, with parameters $\bldgr{\theta}$, that models the imaging process used to obtain the data $\fat{D}$.
This forward model can be ``inverted'' to obtain automated segmentations, as detailed below.

\subsection{Segmentation prior}

We use a segmentation prior based on a deformable probabilistic atlas encoded as a tetrahedral mesh~\citep{VanLeemput2009}. The mesh has node positions $\fat{x}$, governed by a deformation prior distribution defined as:
\begin{align}
p(\fat{x}) \propto \exp \left[ -\mathcal{K} \sum_{m=1}^{M} U_{m}(\fat{x}, \fat{x}_{ref}) \right]
\label{eq:crossPrior}
    ,
\end{align}
where $M$ is the number of tetrahedra in the mesh, $\mathcal{K} > 0$ controls the stiffness of the mesh, and $U_{m}(\fat{x}, \fat{x}_{ref})$ is a topology-preserving cost 
associated with 
deforming the $m^{th}$ tetrahedron from 
its shape in  the atlas's reference position
$\fat{x}_{ref}$~\citep{Ashburner2000}. 

Given a deformed mesh with node positions $\fat{x}$, the probability $p( l_i = k | \fat{x} )$ of observing label $k$ at voxel $i$ is obtained using baricentric interpolation. Assuming conditional independence of the labels between voxels 
finally yields
\begin{align*}
  p(\fat{l}| \fat{x}) 
  = \prod_{i=1}^{I} 
    p( l_i | \fat{x} )
  .
\end{align*}

\subsection{Likelihood function}
\label{sec:crossLikelihood}
We use a multivariate Gaussian intensity model for each of the $K$ different structures, and model the bias field artifact as a linear combination of spatially smooth basis functions that is added to the local voxel intensities~\citep{Wells1996,VanLeemput1999}. Letting $\bldgr{\theta}$ be the collection of the bias field parameters and intensity means and variances, the likelihood function is defined as
\begin{align*}
  p(\fat{D} | \fat{l}, \bldgr{\theta}) = \prod_{i=1}^{I} p(\fat{d}_{i} | l_{i}, \bldgr{\theta}),
  \quad
\end{align*}
\begin{align*}
  p(\fat{d}_{i} | l_i=k, \bldgr{\theta}) 
  = 
  \mathcal{N}( \fat{d}_{i} | \bldgr{\mu}_k + \fat{C}\bldgr{\phi}_{i}, \bldgr{\Sigma}_k ),
\end{align*}
\begin{align*}
  \fat{C} = 
  \left(
    \begin{array}{c}
      \fat{c}_1^T \\
      \vdots \\
      \fat{c}_N^T
    \end{array}
  \right),
  \quad
  \fat{c}_n = 
  \left(
    \begin{array}{c}
      c_{n,1} \\
      \vdots \\
      c_{n,P}
    \end{array}
  \right),
  \quad
  \bldgr{\phi}_i = 
  \left(
    \begin{array}{c}
      \phi_1^i \\
      \vdots \\
      \phi_P^i
    \end{array}
  \right),
\end{align*}
where $P$ denotes the number of bias field basis functions, $\phi_p^i$ is the basis function $p$ evaluated at voxel $i$, and $\fat{c}_n$ collects the bias field coefficients for MRI contrast $n$.
Furthermore, $\bldgr{\mu}_k$ and $\bldgr{\Sigma}_k$ denote the Gaussian mean and variance of structure $k$, respectively.
A flat prior is used for the parameters of the likelihood, i.e., 
$p(\bldgr{\theta}) \propto 1$.

\subsection{Segmentation}

Given an MRI scan $\fat{D}$, 
a corresponding
segmentation is 
obtained by first 
fitting the model to the data:
\begin{align}
    \fat{\hat{x}}, \bldgr{\hat{\theta}} 
    = \argmax_{ 
    \fat{x}, \bldgr{\theta} 
    } p ( \fat{x}, \bldgr{\theta} | \fat{D})
    .
    \label{eq:crossOpt}
\end{align}
The optimization problem in~\eqref{eq:crossOpt} is solved using a coordinate ascent scheme, in which $\fat{x}$ and then $\bldgr{\theta}$ are iteratively updated, each in turn. 
%
    Once the 
    model parameter estimates $\{ \fat{\hat{x}}, \bldgr{\hat{\theta}} \}$
    are available,
    the corresponding maximum a posteriori (MAP) segmentation is obtained as
    \begin{align*}
        \fat{\hat{l}} = \argmax_{ \fat{l} } p ( \fat{l} | \fat{D}, \bldgr{\hat{\theta}} , \fat{\hat{x}} ). 
    \end{align*}
    Since 
    $p( \fat{l} | \fat{D}, \bldgr{\hat{\theta}, \fat{\hat{x}}}) 
    \propto 
    p( \fat{D} | \fat{l}, \bldgr{\hat{\theta}} ) 
    p( \fat{l} | \fat{\hat{x}} )
    $ 
    and therefore
    factorizes over $i$, the optimal segmentation label can be computed 
    for each voxel independently:
    \begin{align}
          \hat{l}_i = \arg \max_k 
    \frac{
    \mathcal{N}( \fat{d}_i | \bldgr{\hat{\mu}}_{k} + \fat{\hat{C}} \bldgr{\phi}_i, \bldgr{\hat{\Sigma}}_{k} ) p( l_i = k | \fat{\hat{x}} ) 
    }{
    \sum_{k'=1}^K \mathcal{N}( \fat{d}_i | \bldgr{\hat{\mu}}_{k'} + \fat{\hat{C}} \bldgr{\phi}_i, \bldgr{\hat{\Sigma}}_{k'} ) p( l_i = k' | \fat{\hat{x}} )
    }
    .
    \label{eq:voxelClassification}
    \end{align}
    More details can be found in~\citep{Puonti2016}. 


\section{Longitudinal method -- SAMSEG-Long}
\label{sec:LongModel}

We now describe how we extend 
SAMSEG
for longitudinal scans. In the remainder of the paper, we call the proposed longitudinal method \textit{SAMSEG-Long}.

In a longitudinal scenario, we aim to compute 
automatic segmentations 
$\{ \fat{l}_t \}_{t=1}^T$ 
from 
$T$ 
consecutive
scans with image intensities 
$\{ \fat{D}_t \}_{t=1}^T$.
%
%
In contrast to the cross-sectional setting where each image is treated
independently, here we can exploit
%
%
%
the fact that all images belong to the same subject to produce more consistent (and potentially more accurate) segmentations.
%
%
Towards this end, we 
introduce 
subject-specific latent variables $\fat{x}_0$ and $\bldgr{\theta}_0$ in the segmentation prior and likelihood function of SAMSEG, respectively.
%
%
%
The purpose of these additional components in the 
model
--
illustrated in blue in Fig.~\ref{fig:model}
--
is to impose a statistical dependency between the time points, encouraging the segmentations to remain similar to one another.
%
%

In the following, we describe how the new subject-specific latent variables are defined in the segmentation prior and likelihood function, 
and
how we 
obtain
the corresponding 
segmentations
accordingly.
We will use the notation $\fat{x}_t$ and $\bldgr{\theta}_t$ to indicate the parameters of the 
prior and likelihood function 
at time $t$,
respectively.

\subsection{Segmentation prior}
\label{subsec:longPrior}

In order to obtain temporal consistency in the segmentation prior, we use the concept of a ``subject-specific atlas''~\citep{Iglesias2016}: a deformation of
the cross-sectional atlas to represent the average subject-specific anatomy across all time points. In particular, we use
$$
  p( \{ \fat{x}_t \}_{t=1}^T | \fat{x}_0 )
  =
  \prod_{t=1}^T  p( \fat{x}_t | \fat{x}_0 )
$$  
with
$$
  p( \fat{x}_{t} | \fat{x}_{0} ) 
  \propto 
  \exp \left[ - \mathcal{K} \sum_{m=1}^{M} U_{m}(\fat{x}_{t}, \fat{x}_{0}) \right]
  ,
$$
where $\fat{x}_0$ are latent atlas node positions encoding subject-specific brain shape, with prior
\begin{align*}
p( \fat{x}_{0} ) \propto \exp \left[ - \mathcal{K}_{0} \sum_{m=1}^{M} U_{m}(\fat{x}_{0}, \fat{x}_{ref}) \right]
.
\end{align*}
Here the mesh stiffness $\mathcal{K}_0$ is a hyperparameter of the model,
the value of which we determine empirically using cross-validation (cf. Sec.~\ref{subsec:Tuning}).
%


Note that this formulation of the longitudinal segmentation prior is very flexible, as it does not impose specific temporal trajectories (e.g., monotonic growth) on the anatomy of the subject.
Furthermore, 
by using 
a very large value for 
its hyperparameter $\mathcal{K}_0$,
$\fat{x}_0$ 
can be forced to remain close to
$\fat{x}_{ref}$,
so that 
cross-sectional segmentation prior of \eqref{eq:crossPrior} 
for each individual time point
is retained as a special case.

%



\subsection{Likelihood function}
\label{subsec:longLikelihood}


In a similar vein, we also introduce subject-specific latent variables to encourage temporal consistency in the Gaussian intensity models. For
each anatomical structure 
$k$,
we condition its Gaussian parameters 
$\{ \bldgr{\mu}_{t,k}, \bldgr{\Sigma}_{t,k} \}_{t=1}^T$
on latent variables 
$\{ \bldgr{\mu}_{0,k}, \bldgr{\Sigma}_{0,k} \}$
using a normal-inverse-Wishart (NIW) distribution:
$$
  p( \{ \bldgr{\theta}_{t} \}_{t=1}^T | \bldgr{\theta}_{0} )
  =
  \prod_{t=1}^T  p( \bldgr{\theta}_{t} | \bldgr{\theta}_{0} )
$$
with
%
\begin{multline*}
   p( \bldgr{\theta}_{t} | \bldgr{\theta}_{0} )
   \propto 
   \prod_{k=1}^K
   \mathcal{N}( \bldgr{\mu}_{t,k} | \bldgr{\mu}_{0,k}, P_{0,k}^{-1} \bldgr{\Sigma}_{t,k} ) 
   \cdot
   \\
   \mathrm{IW}( \bldgr{\Sigma}_{t,k} | P_{0,k} \bldgr{\Sigma}_{0,k}, P_{0,k} - N - 2 )
   ,
\end{multline*} 
where $\bldgr{\theta}_0 = \{ \bldgr{\mu}_{0,k}, \bldgr{\Sigma}_{0,k} \}_{k=1}^K$ collects the latent variables of all structures,
and
is assumed to have a flat prior:
$p(\bldgr{\theta}_0) \propto 1$. 
%
The effect of this longitudinal 
model
is to 
encourage 
the means and variances 
of each structure
to remain similar to 
some ``prototype'' $\bldgr{\mu}_{0,k}$ and $ \bldgr{\Sigma}_{0,k}$, respectively, without having to specify a priori what values these prototypes should take. 
The strength of this effect is governed by 
a
hyperparameter
$P_{0,k} \!\!\ge\!\! 0$ for each structure, which we determine empirically using cross-validation (cf. Sec.~\ref{subsec:Tuning}).


%
%
%
%
%
%
%

Note that no temporal regularization is added to the parameters of the bias field model,
since the bias field 
will typically
vary 
between
MRI sessions.
(Differences in global intensity scaling between time points 
are automatically 
included in
the bias field model as well.)
Furthermore,
by choosing hyperparameters 
$P_{0,k} \!\!=\!\! 0$ 
the temporal regularization of the Gaussian parameters can 
be switched off,
%
%
in which case 
the 
proposed
likelihood function 
devolves into that
of the cross-sectional SAMSEG method 
for each  time point separately
%
(Sec.~\ref{sec:crossLikelihood}).

%
%
%
%
%
%
%
%
%
%

\subsection{Segmentation}
\label{subsec:longSeg}



As in the cross-sectional case, 
segmentations 
are 
obtained by first 
fitting the model to the data:
\begin{align}
  \bldgr{\hat{\theta}}_0, \fat{\hat{x}}_0,
  \{ \fat{\hat{x}}_t, \bldgr{\hat{\theta}}_t \}_{t=1}^T
  = 
  \argmax_{ 
  \bldgr{\theta}_0, \fat{x}_0, 
  \{ \fat{x}_t, \bldgr{\theta}_t \}_{t=1}^T 
} 
  p\Big( 
  \bldgr{\theta}_0, \fat{x}_0,
  \{ \fat{x}_t, \bldgr{\theta}_t \}_{t=1}^T 
  \,
  \big|
  \,
  \{ \fat{D}_t \}_{t=1}^T 
  \Big)
  .
  \label{eq:longOpt}
\end{align}
%
We optimize \eqref{eq:longOpt} with a coordinate ascent scheme, where we iteratively update each variable one at the time. 
Because $p( \fat{x}_t | \fat{x}_0)$ is of the same form as the cross-sectional segmentation prior, and 
the NIW distribution used in
$p( \bldgr{\theta}_t | \bldgr{\theta}_0 ) $ is the conjugate prior for the mean and variance of a Gaussian distribution, estimating 
$\fat{x}_t$ and $\bldgr{\theta}_t$ from $\fat{D}_t$ for given values of 
$\fat{x}_0$ and $\bldgr{\theta}_0$
simply involves performing an optimization of the form of \eqref{eq:crossOpt}  for each time point $t$ separately.
Conversely, for given values $\{ \fat{x}_t, \bldgr{\theta}_t \}_{t=1}^T $ the update for $\bldgr{\theta}_0$ is given in closed form:
%
\begin{align*}
  \bldgr{\mu}_{0,k} 
  & \gets  
  \left(\sum_{t=1}^T \bldgr{\Sigma}_{t,k}^{-1} \right)^{-1} \sum_{t=1}^T \bldgr{\Sigma}_{t,k}^{-1} \bldgr{\mu}_{t,k}
  ,
  \\
    \bldgr{\Sigma}^{-1}_{0,k} 
  & \gets 
  \left(\frac{1}{T} \sum_{t=1}^T \bldgr{\Sigma}_{t,k}^{-1} \right)
  \frac{P_{0,k}}{P_{0,k} - N - 2}
  ,
\end{align*}
%
whereas
updating $\fat{x}_0$ involves
the optimization
\begin{align*}
    \argmin_{\fat{x}_0} \sum_{m=1}^M 
    \left[ 
     \mathcal{K}_{0} U_{m}(\fat{x}_{0}, \fat{x}_{ref}) +
     \mathcal{K} 
     \sum_{t=1}^T
     U_{m}(\fat{x}_{t}, \fat{x}_{0})
    \right]
    ,
\end{align*}
which we solve numerically using a limited-memory BFGS algorithm.

Once all parameters are estimated, we 
obtain segmentations as described in the cross-sectional setting, i.e., by
using \eqref{eq:voxelClassification} for each time point separately.


\begin{figure}[t]
    \centering
    \includegraphics[width=\linewidth]{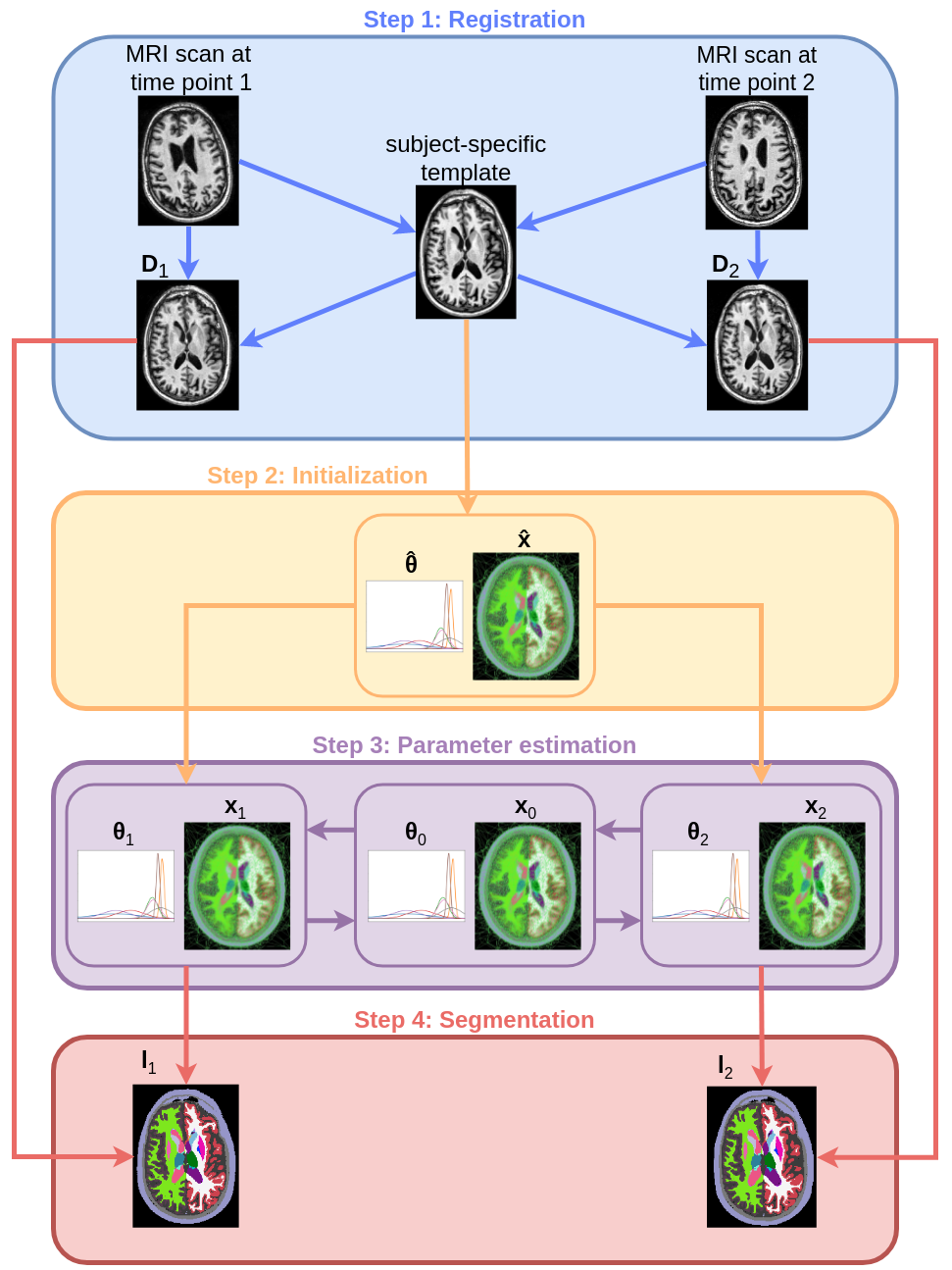} 
    \caption{%
    \revisionTwo{%
    Schematic illustration of the entire longitudinal segmentation process. 
    Although the method is more general, here only the case of single-contrast scans and $T=2$ time points is shown to avoid cluttering. 
    In Step 1, an unbiased subject-specific template is created with an inverse consistent registration method~\citep{Reuter2012}, and each input scan is subsequently registered to it (Sec.~\ref{subsec:Implementation}). 
    The cross-sectional method (Sec.~\ref{sec:CrossModel}) is then applied to the template in Step 2, and the estimated parameters $\hat{\fat{x}}$ and $\hat{\bldgr{\theta}}$ are used to initialize the corresponding parameters $\fat{x}_t$ and $\bldgr{\theta}_t $ at each time point $t$.
    In Step 3, model parameters are estimated in an iterative process (Sec.~\ref{subsec:longSeg}); this involves alternating updates for subject-specific latent variables $\fat{x}_0$ and  $\bldgr{\theta}_0$, with updates for time point parameters $\{ \fat{x}_t, \bldgr{\theta}_t \}_{t=1}^{T}$. 
    Once model parameters estimates are available, in Step 4 each time point is segmented accordingly (Eq.~\eqref{eq:voxelClassification}).
    }
    }
    \label{fig:illustration}
\end{figure}

\subsection{Extension for white matter lesion segmentation}
\label{subsec:LesionModel}


We have previously extended the cross-sectional SAMSEG method using an augmented model that allows it to 
robustly
segment lesions in the white matter~\citep{Cerri2021}.
In this extended version, lesions are modeled with an extra Gaussian in the likelihood function, and with lesion-specific location and shape constraints in the segmentation prior. These 
additional lesion
components are illustrated in gray in the graphical model of Fig.~\ref{fig:model}; we refer the reader to~\citep{Cerri2021} for an in-depth description.

The 
SAMSEG version 
with lesion segmentation ability
can easily be integrated into SAMSEG-Long,
as illustrated in Fig.~\ref{fig:model}.
Due to the highly varying temporal behavior of white matter lesions, we do not explicitly constrain their shape and appearance over time, as this could potentially degrade the segmentation performance of the method. 
As a result, there are no direct dependencies between the
lesion-specific components of the model and the latent variables $\fat{x}_0$ and $\bldgr{\theta}_0$ (note the absence of direct arrows between the two sets of variables in Fig.~\ref{fig:model}). 
Computing longitudinal whole-brain and white matter lesion segmentations can therefore follow the same procedure described before (cf. Sec.~\ref{subsec:longSeg}), with only a few modifications in how, for each time point $t$, the 
parameters 
$\fat{x}_t$ and $\bldgr{\theta}_t$
are estimated, and individual segmentations $\fat{l}_t$ are computed~\citep{Cerri2021}.

\section{Implementation}
\label{subsec:Implementation}

In our current implementation, it is assumed that all time points have been registered and resampled to the same image grid prior to segmentation.
%
To
avoid 
introducing
spurious biases 
by
not treating
all time points in exactly the same way
(e.g., by resampling follow-up scans to a baseline scan)~\citep{Reuter2011}, 
for this purpose
we 
use
an unbiased within-subject template created with an inverse consistent registration method~\citep{Reuter2012}.
This template is a robust representation of the average subject
anatomy over time, and we use it as an unbiased reference to register all time points to, resulting in resampled images 
that then form the input to our segmentation algorithm.
In case of multi-contrast images, 
this procedure is performed for one specific contrast 
(T1w in the experiments used in this paper),
and the remaining contrasts 
(FLAIR in the experiments)
are subsequently registered and resampled to the first contrast for each time point individually.%

To initialize the proposed algorithm,
we first apply the cross-sectional method to the unbiased template,
and use the estimated model parameters 
$\fat{\hat{x}}$ and $\bldgr{\hat{\theta}}$ 
to initialize the corresponding parameters
$\fat{x}_t$ and $\bldgr{\theta}_t$ at each time point $t$.
The model fitting procedure
of \eqref{eq:longOpt},
%
which interleaves updating the latent variables 
$\{ \fat{x}_0, \bldgr{\theta}_0 \}$
with updating the parameters $\{ \fat{x}_t, \bldgr{\theta}_t \}_{t=1}^T$, is then run for five iterations, which we
have found to be sufficient to reach convergence.

\revisionTwo{An illustration of the entire longitudinal segmentation process is provided in Fig.~\ref{fig:illustration}.}
Our implementation builds upon the C++ and Python code of~\citep{Puonti2016} and~\citep{Cerri2021},
and is publicly available from FreeSurfer\footnote{\url{https://surfer.nmr.mgh.harvard.edu/fswiki/Samseg}}. 
Segmenting one subject with 1mm$^3$ isotropic resolution \revision{and image size of $256^3$} takes approximately 10 minutes per time point for SAMSEG-Long, while 5 additional minutes per time point are needed when segmenting also white matter lesions (measured on an Intel 12-core i7-8700K processor).

\revisionTwo{\subsection{Hyperparameter tuning}}
\label{subsec:Tuning}

SAMSEG-Long has hyperparameters $\mathcal{K}_0$ and $P_{0,k}$ that control, respectively, the strength of the regularization in the segmentation prior and likelihood function.
Good choices for the value of these hyperparameters will aim to
minimize differences between scans acquired within a short interval of time,
while simultaneously maximizing the ability to detect known atrophy trajectories in different patient groups.

We therefore
tuned these hyperparameters by applying a grid search 
\revisionTwo{%
using
80 
test-retest scans 
and
80 
longitudinal scans of both cognitively normal (CN) and AD subjects.%
}
In particular, 
\revisionTwo{%
using 
T1w 
images
from 
the MIRIAD-TR-HT (CN=10, AD=30) and the ADNI-HT (CN=37, AD=53) datasets 
summarized in Table~\ref{tab:datasets} and detailed in~\ref{app:datasets},%
}
%
%
%
we searched from the following values of the hyperparameters:
$\mathcal{K}_0= \{ 5 \mathcal{K}, 10 \mathcal{K}, 14 \mathcal{K}, 15 \mathcal{K}, 20 \mathcal{K} \}$
and 
$P_{0,k} = \{ 0.25 N_k, 0.5 N_k, 0.75 N_k, N_k, 1.25 N_k \}$,
where $N_k$ is the number of voxels assigned to class $k$ in the cross-sectional segmentation of the within-subject template. 
%
%

\begin{figure}[t]
    \centering
    \includegraphics[scale=0.245]{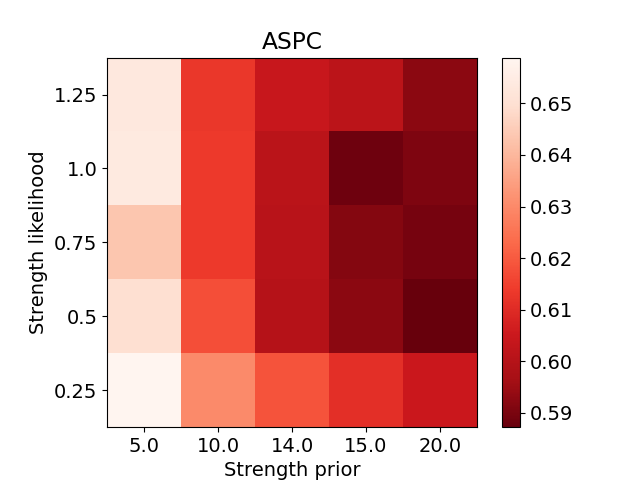} 
    \includegraphics[scale=0.245]{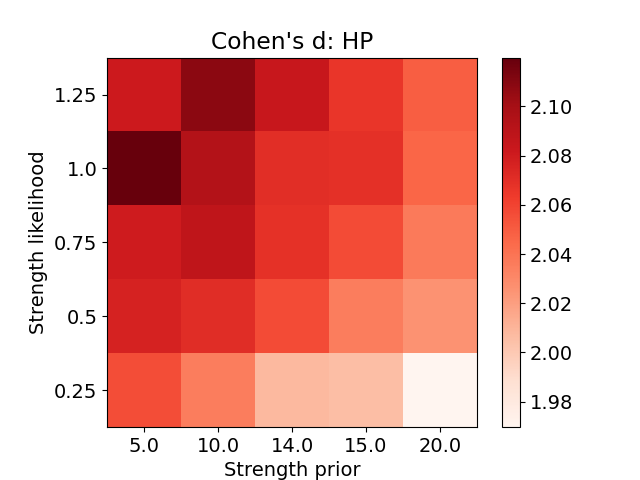} \\
    \includegraphics[scale=0.245]{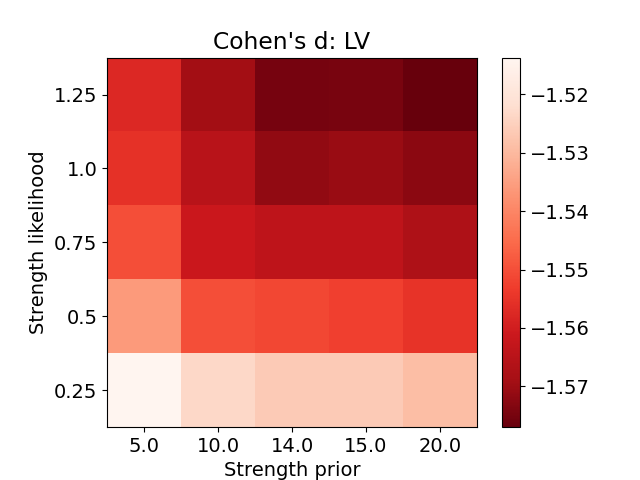}
    \includegraphics[scale=0.245]{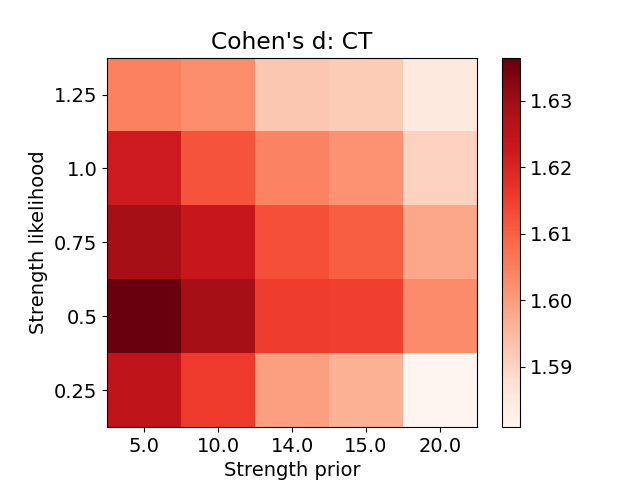}
    \caption{Hyperparameter tuning by grid search over the hyperparameters of the proposed method. Top left: ASPC values across subjects and structures for the 80 test-retest T1w scans of the 40 subjects of the MIRIAD-TR-HT dataset (CN=10, AD=30). Top right and bottom row: Cohen's effect sizes computed from APCs estimates of the 80 subjects of the ADNI-HT dataset (CN=37, AD=53) for Hippocampus (HP), Lateral Ventricles (LV) and Cerebral Cortex (CT). 
    \revision{APSC=Absolute Symmetrized Percent Change, CN=Cognitive Normal, AD=Alzheimer's Disease.}.
    }
    \label{fig:hyperparameterTuning}
\end{figure}

The results are summarized in Fig.~\ref{fig:hyperparameterTuning}, which shows
ASPC values in the test-retest scenario, as well as Cohen's d effect sizes 
of APC values
between CN and AD patients for hippocampus, lateral ventricles and cerebral cortex -- three structures known to be strongly affected in AD~\citep{Lombardi2020}.
\revisionTwo{%
The ASPC and APC metrics are defined in detail in Sec.~\ref{sec:metricsAndMeasures},
but in short 
assess
volumetric changes between test and retest scans (ASPC),
and 
the 
yearly rate of volume changes in longitudinal scans
(APC), 
both expressed as a percentage.
}

%
%

The proposed method yielded consistent performance overall, with only minor differences between the various hyperparameter value combinations: 
ASPC 
and
Cohen's d
values 
varied within a 5.8\% and 3.7\% range compared to the average performance, respectively.
Nevertheless, for the purpose of having fixed values for these hyperparameters, we used the combination $\mathcal{K}_0=20\mathcal{K}$ and $P_{0,k} = 0.5N_k, \forall k$ for all the 
experiments \revisionTwo{described below}.

\vspace{0.5cm}

\section{Experiments}
\label{sec:experiments}

In order to evaluate the performance of SAMSEG-Long, we conducted experiments on multiple datasets acquired with many different scanner platforms, field strengths, acquisition protocols, and image resolutions.
These datasets 
contain images of cognitively normal 
subjects as well as AD and MS patients,
and 
differ 
both
in
the
number and timing of their longitudinal follow up scans, 
as well as in the number of MRI contrasts that are acquired.
%
%
A summary of the datasets can be found in Table~\ref{tab:datasets}, with more detailed information for each individual dataset in~\ref{app:datasets}.
Taken together, we believe these datasets are an excellent source to demonstrate the robustness and generalizability of the proposed longitudinal method, which does not need to be retrained or tuned on any of these datasets. 

As a first experiment,
%
we
evaluated the method's test-retest reliability.
Since test-retest scans are acquired within a minimal interval of time \revision{(usually within the same scan session or within a couple of weeks)}, no biological variations in the various structures are expected, and we therefore evaluated the ability of the method to produce consistent segmentations in these settings.
Although this property is essential in a longitudinal segmentation method, an algorithm that produces the same segmentation for each given time point would, by definition, also have perfect test-retest reliability.
Additional experiments are therefore needed to evaluate the 
ability of the method to also detect real longitudinal 
changes
if they exist. 
%
%
%
%
%
Ideally, this would involve comparing the longitudinal segmentations computed by the proposed method with manually delineated longitudinal data. 
However, to the best of our knowledge, such ground truth data is not 
currently
available.
We therefore performed an \emph{indirect} evaluation of the 
sensitivity of the
method, by assessing its ability to 
detect
known 
differences
between 
the temporal trajectories 
in 
different
patient groups (CN vs.~AD,  stable vs.~progressive MS).



For all our experiments, we report the performance of the ``vanilla'' SAMSEG-Long method as described in Sec.~\ref{subsec:longSeg} -- except for MS patients, for which we use the method with its white matter lesion segmentation extension (Sec.~\ref{subsec:LesionModel}).

\begin{table*}[!ht]
\resizebox{1\textwidth}{!}{%
\begin{tabular}{|c|c|c|c|c|c|c|c|}
\hline
\multicolumn{1}{|c|}{Experiment} & \multicolumn{1}{c|}{Dataset} & Subjects & \# scans & avg-tp (min, max) & time-tp (min, max) & \multicolumn{1}{c|}{Scanner} & \multicolumn{1}{c|}{Sequence} \\ \hline
\multirow{2}{*}{Hyperparameter tuning} 
& MIRIAD-TR-HT & CN=10, AD=30 & 80 & 2 (2, 2) & 0 (0, 0) & GE Signa 1.5T & IR-FSPGR \\ \cline{2-8} 
& ADNI-HT & CN=37, AD=53 & 285 & 3.56 (2,5) & 313 (107, 1121) & Multiple 3T scanners & \begin{tabular}[c]{@{}l@{}}MP-RAGE \\ IR-FSPGR\end{tabular} \\ \hline
\multirow{6}{*}{Test-retest reliability} 
& MIRIAD-TR & CN=13 , AD=16 & 146 & 2 (2, 2) & 0 (0, 0) & GE Signa 1.5T & IR-FSPGR \\ \cline{2-8}
& OASIS-TR & \begin{tabular}[c]{@{}c@{}}CN=72 , CV=14, \\ AD=64\end{tabular} & 1845 & 3.55 (3, 4) & 0 (0, 0) & Siemens Vision 1.5T & MP-RAGE \\ \cline{2-8} 
& Munich-TR & MS=2 & 34 & 5.67 (5, 6) & 3 (2, 7) & \begin{tabular}[c]{@{}l@{}}Philips Achieva 3T\\ Siemens Verio 3T\\ GE Signa MR750 3T\end{tabular} & \begin{tabular}[c]{@{}l@{}}IR-FSPGR \\ MP-RAGE \\ FLAIR\end{tabular} \\ \cline{1-8} 
\multirow{2}{*}{Detecting disease effects} 
& ADNI & CN=66, AD=64 & 477 & 3.70 (2, 5) & 298 (65, 903) & \begin{tabular}[c]{@{}l@{}}Multiple 1.5T and \\ 3T scanners\end{tabular} & \begin{tabular}[c]{@{}l@{}}IR-FSPGR \\ MP-RAGE\end{tabular} \\ \cline{2-8} 
& OASIS & CN=72, AD=64 & 336 & 2.47 (2, 5) & 702 (182, 1510) & Siemens Vision 1.5T & MP-RAGE \\ \cline{2-8}  
& Munich & \begin{tabular}[c]{@{}c@{}} MS=200 \\ (S-MS=100, P-MS=100) \end{tabular} & 1289 & 6.45 (2, 24) & 353 (18, 3287) & Philips Achieva 3T & \begin{tabular}[c]{@{}l@{}}MP-RAGE\\ FLAIR\end{tabular} \\ \hline
\revision{\begin{tabular}[c]{@{}c@{}}Longitudinal lesion \\ segmentation\end{tabular}} & \revision{ISBI} & \revision{MS=14} & \revision{61} & \revision{4.36 (4, 6)} & \revision{391 (299, 503)} & \revision{Philips 3T} & \begin{tabular}[c]{@{}l@{}}\revision{MP-RAGE}\\ \revision{FLAIR}\end{tabular}  \\ \hline
\end{tabular}
}
\caption{
Summary of the experiments and datasets used in the paper. CN=Cognitive Normal, CV=Converted, AD=Alzheimer's Disease, MS=Multiple Sclerosis, S-MS=Stable-MS, P-MS=Progressive-MS, \# scans=total number of scans, tp=time points, avg-tp=average number of time points per subject, time-tp=average time in days between each time point, HT=Hyperparameter Tuning, TR=Test-Retest, IR-FSPGR=Inversion Recovery prepared - Fast SPoiled Gradient Recalled, MP-RAGE=Magnetization Prepared - RApid Gradient Echo, FLAIR=FLuid Attenuation Inversion Recovery.
For more details about each individual dataset, see~\ref{app:datasets}.
}
\label{tab:datasets}
\end{table*}

\subsection{Benchmark methods}

In order to benchmark the 
longitudinal
whole-brain segmentation performance of SAMSEG-Long, 
we compared 
it against that of
SAMSEG (which is cross-sectional), and the longitudinal stream of FreeSurfer 7.2~\citep{Reuter2012}, called Aseg-Long in the remainder of the paper.
Aseg-Long is the only publicly available and extensively validated longitudinal method that segments the same neuroanatomical structures as our method, representing a natural benchmark for evaluating its whole-brain segmentation performance.
\revision{%
However, other 
tools exist that 
have reported
better performance for estimating longitudinal volume changes in specific structures, such as the hippocampus, lateral ventricles or gray matter~\citep{Nakamura2014,Guizard2015}.
}
%
%
%
Note that Aseg-Long is unable to process multi-contrast scans, hence no comparison was performed on such data.

For evaluating the lesion segmentation component of SAMSEG-Long, 
we compared 
its performance 
against that of both SAMSEG and the longitudinal white matter lesion segmentation method of~\citep{Schmidt2019}, called LST-Long in the remainder of the paper.
LST-Long is one the few methods that have a publicly available implementation.
It 
segments 
lesions 
from T1-weighted and FLAIR MRI scans
with
multiple time points,
and does not require retraining when tested on unseen data.
Unlike our method, however, it 
does not provide further segmentations of the various neuroanatomical structures
beyond white matter lesions.
%



\subsection{Metrics and measures}
\label{sec:metricsAndMeasures}


Although the proposed method segments more structures,
we concentrated on the following 25 main neuroanatomical regions: left and right cerebral white matter (WM), cerebellum white matter (CWM), cerebellum cortex (CCT), cerebral cortex (CT), lateral ventricle (LV), hippocampus (HP), thalamus (TH), putamen (PU), pallidum (PA), caudate (CA), amygdala (AM), nucleus accumbens (AC) and brain stem (BS). 
To avoid cluttering, we merged the results between right and left structures.
For experiments that include MS patients, white matter lesion (LES) results are also reported.

For evaluating test-retest reliability, we computed the Absolute Symmetrized Percent Change (ASPC) for each structure, defined as
\begin{align*}
    \text{ASPC} 
    = \frac{100 \cdot | \revision{v_2 - v_1} |}{(v_1 + v_2) / 2 },
\end{align*}
where $v_1$ and $v_2$ represent the volume of the structure in scan 1 and scan 2, respectively. 
\revision{To check for under- and over-segmentation trends, we also computed Symmetrized Percent Change (SPC), defined in the same way as ASPC but without the absolute value.} 

For assessing 
a method's ability to detect disease effects,
we computed the Annualized Percentage Change (APC):
We fitted a line to the subject's volumetric measurements,
\revision{plotted as a function of time from the baseline,}
and computed the APC as the ratio of its slope to its intercept (evaluated at the time of the first scan).
\revision{%
A negative APC value thus corresponds to a yearly shrinkage, in percentage, of the structure of interest, while a positive APC value indicates a yearly growth.
}%
Effect sizes between APCs of two patient groups were then computed using Cohen's d~\citep{Cohen2013}.
\revision{
We also report the minimum number of subjects needed to detect a statistically significant difference in atrophy rates between two patient groups, by first fitting a generalized linear model with APCs and corresponding patient groups. We then performed a power analysis using the computed model coefficients and noise variance, with 80\% power and 0.05 significance level~\citep{Cohen2013}.%
}

%
For 
evaluating sensitivity to lesion activity in MS patients,
we additionally assessed the 
apparent speed of lesion volume growth and shrinkage
between segmentations at consecutive time points.
Specifically,
we computed annualized lesion volume \emph{increase} 
(LES\_I) 
and \emph{decrease} 
(LES\_D) 
as
\begin{align*}
    \text{LES\_I} =& 
    \frac{1}{T-1}\sum_{t=1}^{T-1} \frac{Nz_{t, t+1}}{\Delta_{t,t+1}}
\end{align*}
and
\begin{align*}
    \text{LES\_D} =& 
    \frac{1}{T-1}\sum_{t=1}^{T-1} \frac{{Nz_{t+1, t}}}{\Delta_{t,t+1}}
    ,
\end{align*}
respectively.
Here 
$Nz_{t,t+1}$ 
counts the number of voxels that were labeled as lesion at time point 
$t+1$
but not at time point 
$t$,
while 
$\Delta_{t, t+1}$ 
is the time passed between scan 
$t$ 
and scan 
$t+1$.
Intuitively, LES\_I (LES\_D) counts by how many voxels the total lesion volume has grown (shrunk) in a year, on average, \revision{\emph{ignoring potential lesion areas where there was 
simultaneous
shrinkage (growth)}}. 
\revision{%
In MS, lesions
\revisionTwo{as depicted by MRI}
not only appear and grow~\citep{Gaitan2011} but also shrink and disappear\revisionTwo{~\citep{Sethi2017,Battaglini2022}}, 
and these metrics
have therefore been suggested as 
markers of disease activity~\citep{Pongratz2019}.
%
%
%
%
%
%
}


\revision{To evaluate the lesion segmentation performance of the proposed method against ground truth delineations, we computed Dice coefficients as:
\begin{align*}
    \text{DICE}_{X, Y} = \frac{2 |X \cap Y|}{|X| + |Y|},
\end{align*}
where $X$ and $Y$ denote segmentation masks, and $| \cdot |$ counts the number
of voxels in a mask.
}
\section{Results}
\label{sec:results}



Throughout this section, we will refer to specific datasets with their names as defined in Table~\ref{tab:datasets}. As a mnemonic, 
datasets used to evaluate \emph{test-retest} reliability 
have an affix ``-TR''
in their names.
%
%
Whenever boxpots are used, the median is indicated by a horizontal line, plotted inside boxes that extend from the first to the third quartile values of the data. The range of the data is indicated by whiskers extending from the boxes, with outliers represented by ``x'' symbols.

\subsection{Test-retest reliability}
\label{subsec:TestRetest}

\begin{figure*}[!ht]
    \centering
    \begin{minipage}[t]{.99\linewidth}
        \includegraphics[height=6.5cm]{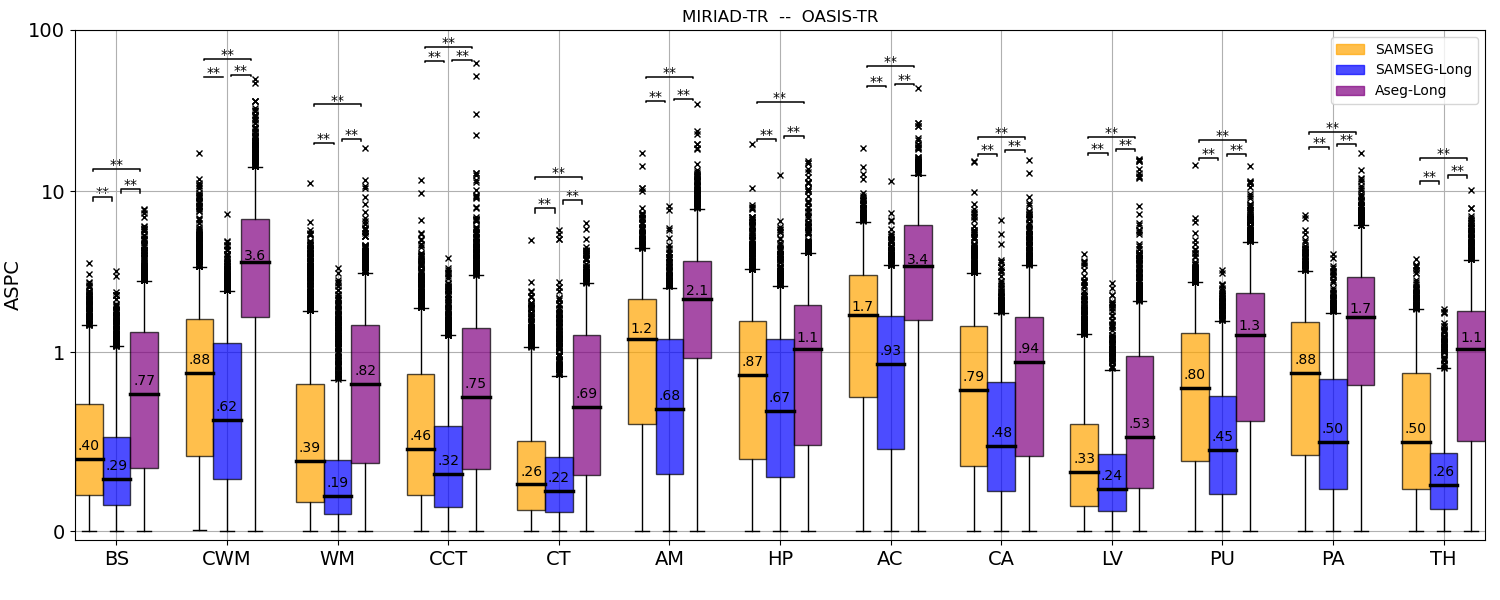}
    \end{minipage}
    \begin{minipage}[t]{.99\linewidth}
        \includegraphics[height=6.5cm]{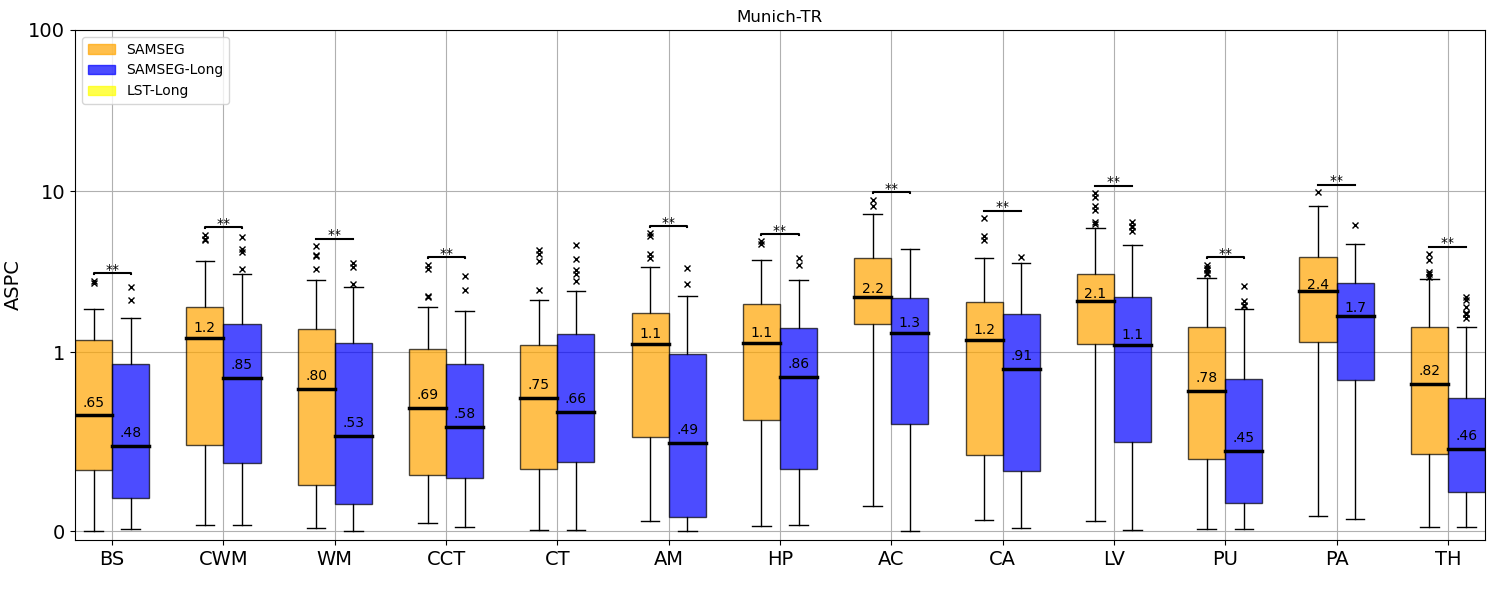}
        \hspace{-0.32cm}
        \includegraphics[height=6.5cm]{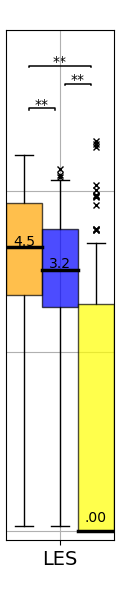}
    \end{minipage}
    \caption{
    ASPC values for the test-retest T1w scans of the combined MIRIAD-TR (\# scans: 146, AD=13, CN=16) and OASIS-TR (\# scans: 1845, AD=72, Converted=14, CN=64) datasets (top) and for the test-retest T1w-FLAIR scans of the Munich-TR (\# scans: 34, MS=2) dataset (bottom). 
    \revision{The yellow boxplot shows the result for LST-Long for white matter lesions.}
    ASPC values were computed for different brain structures and white matter lesion (LES) for the proposed method and the benchmark methods.
    Within each boxplot, the median \revision{ASPC} value is also reported. 
    Statistically significant differences between two methods were computed with a Wilcoxon signed-rank test, and are indicated by asterisks \revision{(``**'' indicates p-value $<$ 0.01).}
    \revision{ASPC=Absolute Symmetrized Percent Change, CN=Cognitive Normal, AD=Alzheimer's Disease, T1w=T1-weighted, FLAIR=FLuid Attenuation Inversion Recovery, MS=Multiple Sclerosis,}
    \revision{BS=brain stem, CWM=cerebellum white matter, WM=cerebral white matter, CCT=cerebellum cortex, CT=cerebral cortex, AM=amygdala, HP=hippocampus, AC=nucleus accumbens, CA=caudate, LV=lateral ventricle, PU=putamen, PA=pallidum, TH=thalamus, LES=white matter lesion.}
    }
    \label{fig:ASPCs}
\end{figure*}

In order to evaluate if the proposed longitudinal method produces consistent segmentations over time, we assessed its performance on test-retest scans of three different datasets: 146 T1w test-retest scans of the 29 subjects of the MIRIAD-TR dataset (CN=13, AD=16), 1845 T1w test-retest scans of the 150 subjects of the OASIS-TR dataset (CN=72, Converted=14, AD=64) and 34 T1w and FLAIR test-retest scans of the 2 MS patients of the Munich-TR dataset. 
We thus computed automated segmentations for SAMSEG-Long and the benchmark methods, and assessed test-retest reliability performance in terms of ASPC values.
When more than two test-retest scans were available for a subject, APSC values were computed for each possible combination of test-retest scan pairs. 
\revision{The results are shown in Fig.~\ref{fig:ASPCs}.}

Since we observed similar results in the MIRIAD-TR and the OASIS-TR datasets, 
we here only report on their combined results. 
(We redirect the reader to Fig.~\ref{fig:ASPCsMIRIAD_OASIS} for the results on the individual datasets.)
On the combined MIRIAD-TR/OASIS-TR dataset,  
the median ASPC across all structures was 
best for  SAMSEG-Long: 0.39,
compared to 0.62 for SAMSEG and 1.12 for Aseg-Long. 
Similarly, on the Munich-TR dataset the median ASPC was 0.69 for SAMSEG-Long vs.~1.07 for SAMSEG (Aseg-Long cannot be run on this multi-contrasts dataset).
The overall weaker performance on the 
Munich-TR scans 
can be explained by the fact that
these were acquired  within a 3-week interval,
whereas the scans of the combined MIRIAD-TR/OASIS-TR dataset
were acquired within a single scan session without repositioning.
\revision{
Fig.~\ref{fig:ASPCs} top shows
a high number of outliers for the combined MIRIAD-TR/OASIS-TR dataset, with large ASPC values especially for Aseg-Long. This is mostly due to the high number of test-retest scans (approximately 2,000) of this combined dataset.}




%


As for ASPC values of white matter lesions in the Munich-TR dataset
\revision{(Fig.~\ref{fig:ASPCs} bottom right)}%
, SAMSEG-Long outperformed SAMSEG (median ASPC: 4.5 vs.~3.2) even though the method does not explicitly regularize white matter lesions longitudinally (see Sec.~\ref{subsec:LesionModel}). 
This may indicate that more consistent model parameter estimates were obtained for SAMSEG-Long compared to SAMSEG as a result of enforcing temporal consistency on all the other structures.
We also observe almost perfect test-retest reliability performance for LST-Long (median ASPC: 0.0), 
thereby outperforming
both SAMSEG-LONG and SAMSEG by a large margin in this regard (but see below). 

\revision{
To check whether 
some of the 
methods are prone to under- or over-segmenting 
on test-retest scans, we also 
report the SPC values (i.e., without taking absolute values) on the same data.
The results, shown in Fig.~\ref{fig:SPCs}, 
did not indicate any particular trend in under- or over-segmenting specific structures for any of the methods
(median SPC values for all the methods are close to 0), with SAMSEG-Long having the smallest SPC variances, followed by SAMSEG and Aseg-Long. 
}

\subsection{Detecting disease effects}

\begin{figure*}[!ht]
    \centering
    \begin{minipage}[t]{.95\linewidth}
        \includegraphics[width=\textwidth]{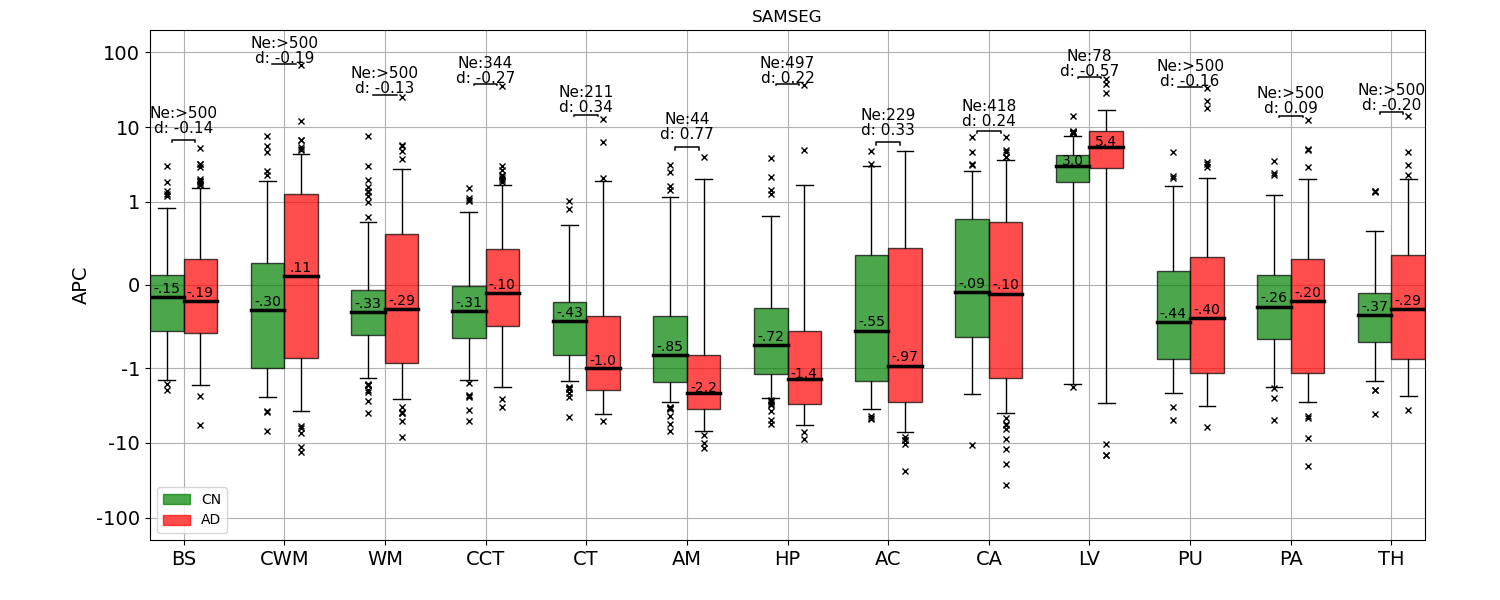}
    \end{minipage}
    \begin{minipage}[t]{.95\linewidth}
        \includegraphics[width=\textwidth]{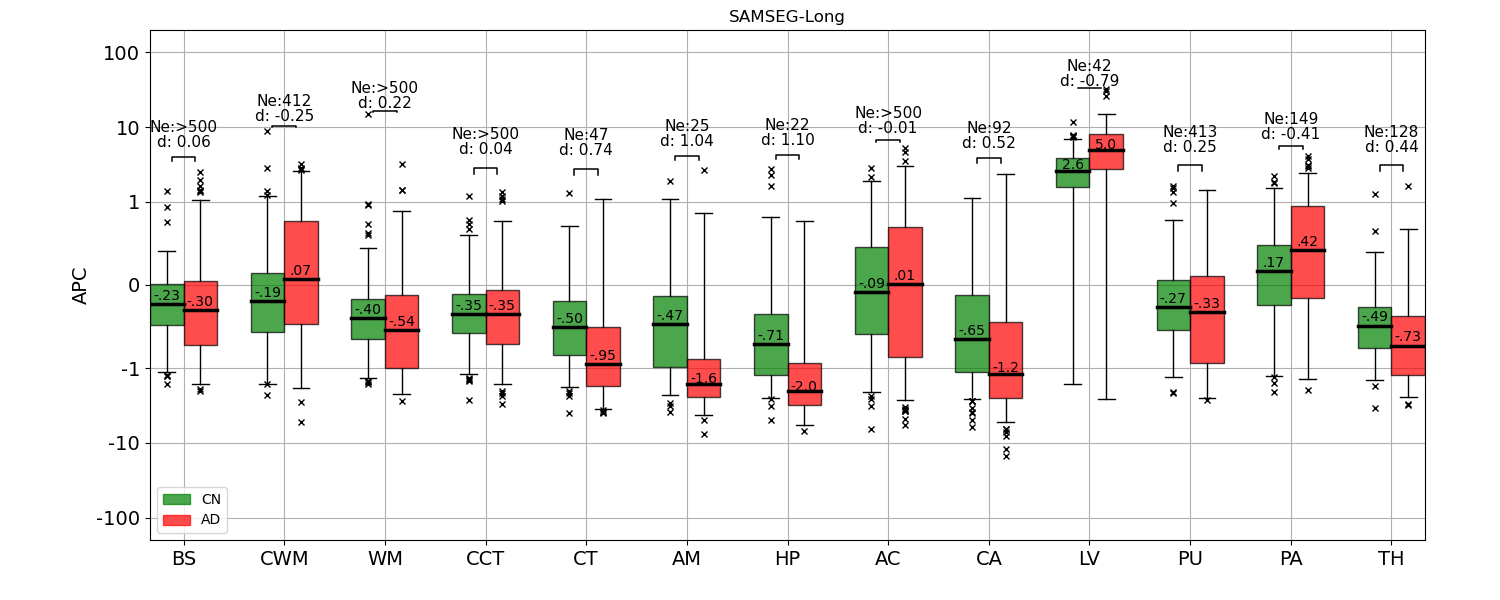}
    \end{minipage}
        \begin{minipage}[t]{.95\linewidth}
        \includegraphics[width=\textwidth]{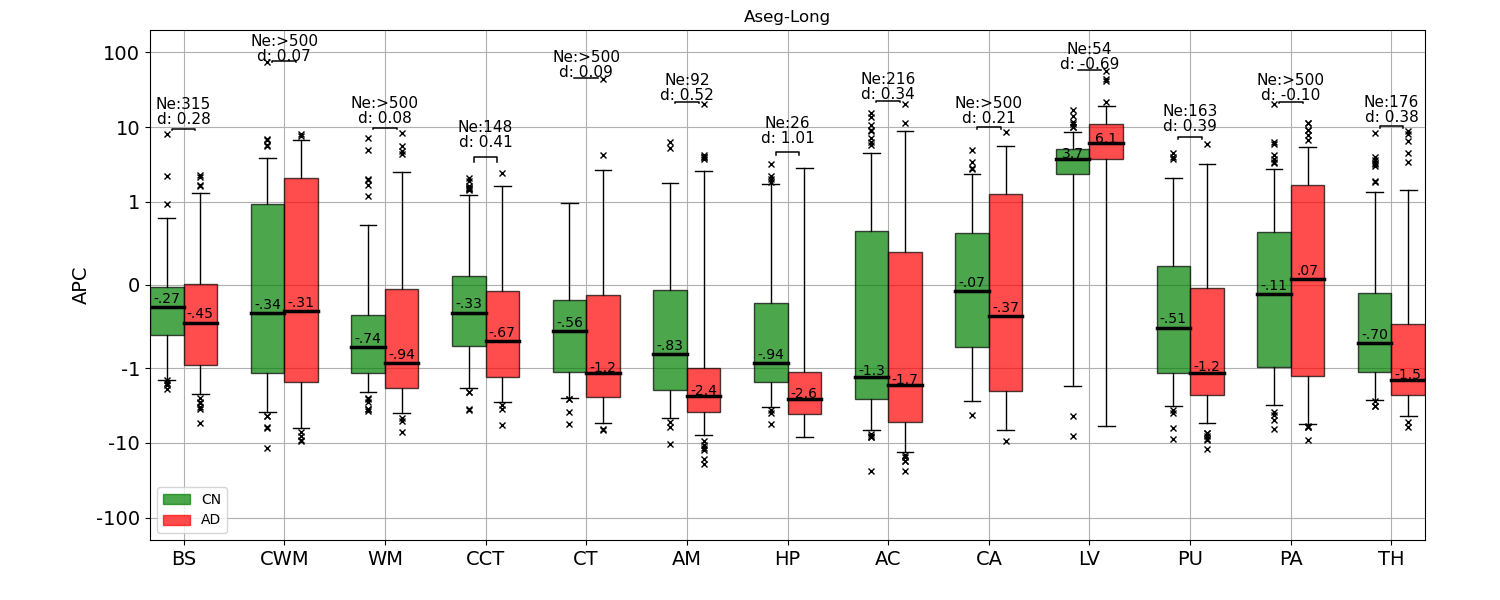}
    \end{minipage}
    \caption{
    APCs computed from the T1w scans of the 130 subjects of the ADNI dataset (CN=66, AD=64) and \revisionTwo{136} subjects of the OASIS dataset (CN=72, AD=64) for SAMSEG, SAMSEG-Long, and Aseg-Long.
    Cohen's d effect size\revision{ (d) and effective number of subjects (Ne) computed from a power analysis (80\% power, 0.05 significance level) are} reported above each pair of box plots.
    Within each boxplot, the median \revision{APC} 
    value is also indicated.
    \revision{APC=Annualized Percentage Change, CN=Cognitive Normal, AD=Alzheimer's Disease, T1w=T1-weighted,}
    \revision{BS=brain stem, CWM=cerebellum white matter, WM=cerebral white matter, CCT=cerebellum cortex, CT=cerebral cortex, AM=amygdala, HP=hippocampus, AC=nucleus accumbens, CA=caudate, LV=lateral ventricle, PU=putamen, PA=pallidum, TH=thalamus.}
    }
    \label{fig:APCs_AD}
\end{figure*}

\iffalse
Finally, we assessed the ability of the various methods to detect longitudinal disease effects. 

For each method we computed APC values for all the structures, and analyzed how they differ between CN and AD patients for the ADNI (CN=66, AD=64) and OASIS (CN=72, AD=64) datasets.

\else
\revision{We assessed the ability of the various methods to detect longitudinal disease effects by computing APC values for all the structures and analyzing how they differ between CN and AD patients for the ADNI (CN=66, AD=64) and OASIS (CN=72, AD=64) datasets.}
\fi
%
%
%
Since we found similar findings across the two datasets, 
Fig.~\ref{fig:APCs_AD}
shows the results for both datasets combined
to ease readability.
(We redirect the reader to Fig.~\ref{fig:APCs_ADNI} and Fig.~\ref{fig:APCs_OASIS} for individual dataset results.)
All the methods were able to capture well-known differences in the atrophy trajectories of the hippocampus, amygdala and lateral ventricles between CN and AD patients~\citep{Lombardi2020}. Effect sizes in these three structures
differed between methods, with SAMSEG-Long having higher values
for Cohen's d [0.79-1.10], followed by Aseg-Long [0.52-1.01] and SAMSEG [0.22-0.77].
\revision{%
The results of the power analysis closely mimick these findings: SAMSEG-Long requires fewer subjects to detect the differences in APC [22-42] compared to Aseg-Long [26-54] and SAMSEG [44-497].
}%
Interestingly, only SAMSEG-Long showed a strong effect size for cerebral cortex (Cohen's d: 0.74), whose atrophy trajectories are known to differ between CN and AD patients~\citep{Edmonds2020}, while both Aseg-Long and SAMSEG report lower effect sizes (Aseg-Long: 0.09, SAMSEG: 0.34).

\begin{figure*}[!ht]
    \centering
    \begin{minipage}[t]{.95\linewidth}
        \includegraphics[width=\textwidth]{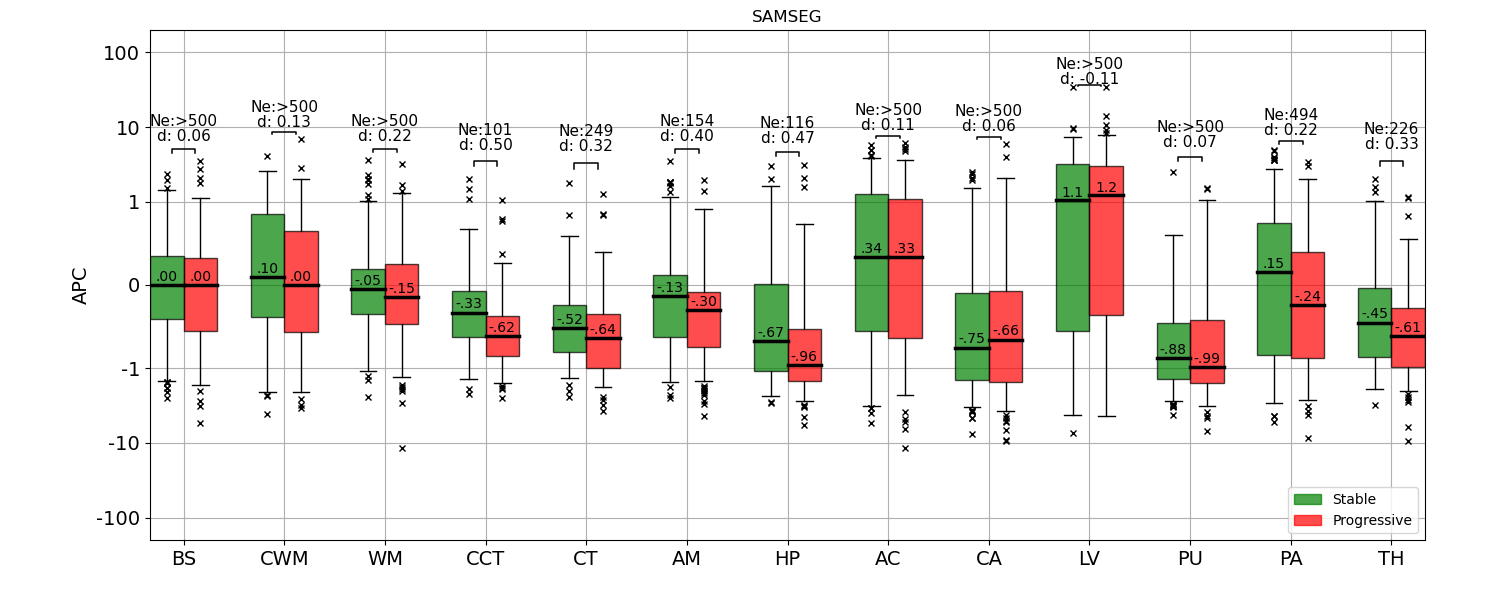}
    \end{minipage}
    \begin{minipage}[t]{.95\linewidth}
        \includegraphics[width=\textwidth]{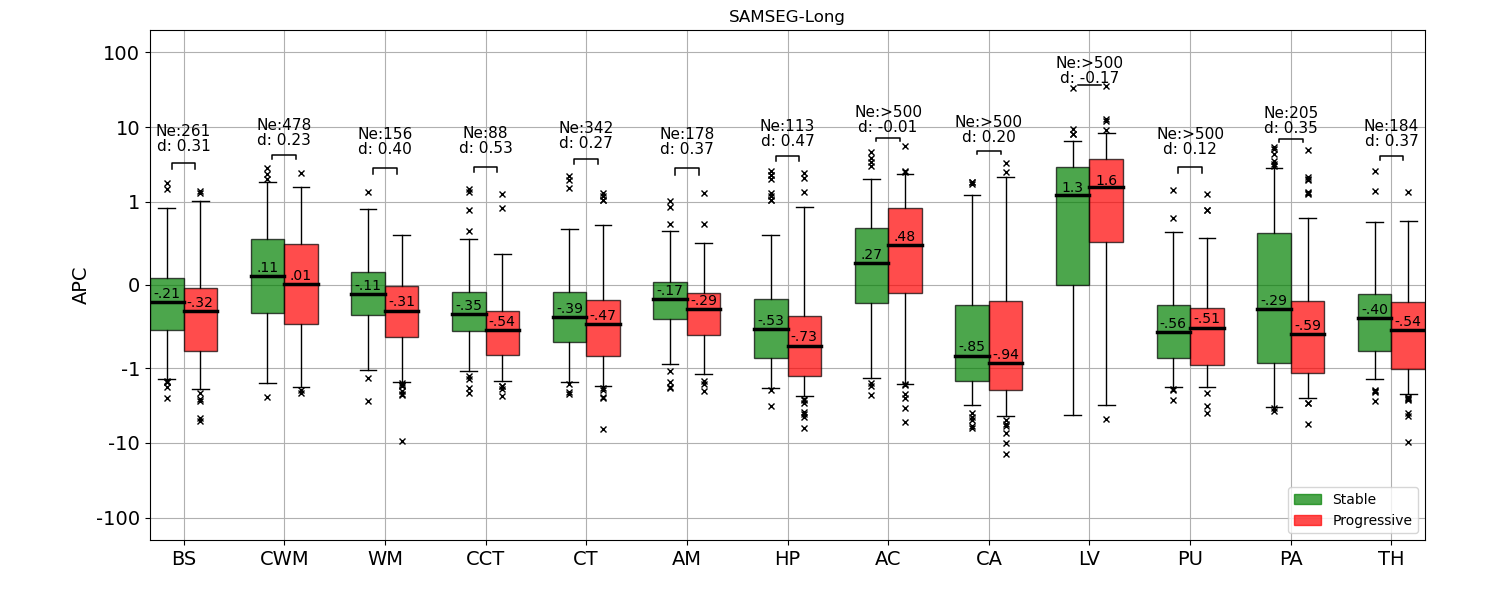}
    \end{minipage}
    \caption{
    APCs of several structures computed from the T1w and FLAIR scans of the 200 patients of the Munich dataset (100 stable MS, and 100 progressive MS) for SAMSEG and SAMSEG-Long.
    For each comparison, Cohen's d effect size\revision{ (d) and effective number of subjects (Ne) computed from a power analysis (80\% power, 0.05 significance level) are} shown above the boxplots.
    Within each boxplot, the median \revision{APC} value is also reported.
    \revision{APC=Annualized Percentage Change, T1w=T1-weighted, FLAIR=FLuid Attenuation Inversion Recovery, MS=Multiple Sclerosis,}
    \revision{BS=brain stem, CWM=cerebellum white matter, WM=cerebral white matter, CCT=cerebellum cortex, CT=cerebral cortex, AM=amygdala, HP=hippocampus, AC=nucleus accumbens, CA=caudate, LV=lateral ventricle, PU=putamen, PA=pallidum, TH=thalamus.}
    }
    \label{fig:APCs_MS}
\end{figure*}

\begin{figure*}[!ht]
    \centering
    \includegraphics[width=0.46\linewidth]{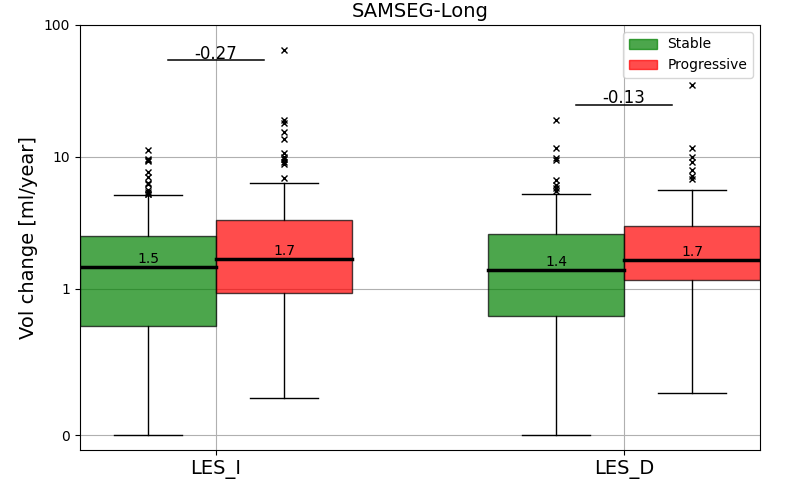}
    \includegraphics[width=0.46\linewidth]{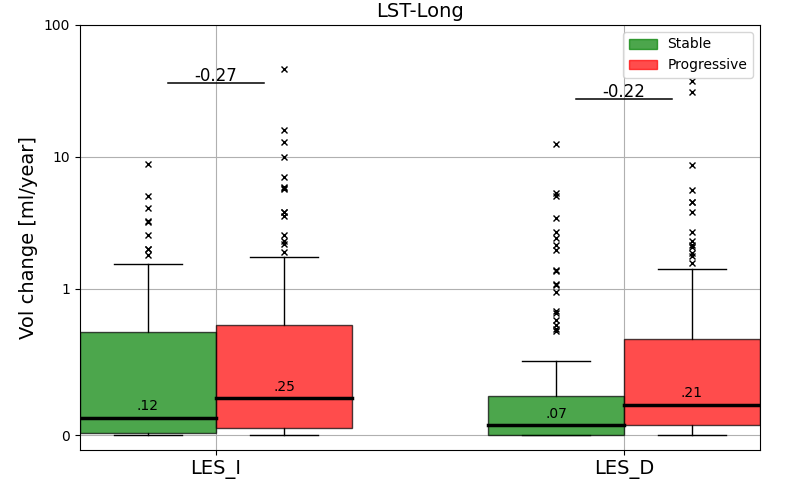}
    \caption{
    Lesion volume increase (LES\_I) and decrease (LES\_D) computed from the T1w and FLAIR scans of the 200 patients of the Munich dataset (100 stable MS, and 100 progressive MS) for SAMSEG-Long and LST-Long. 
    For each comparison, Cohen's d effect size is shown above the boxplots.
    Within each boxplot, the median value is also reported.
    \revision{T1w=T1-weighted, FLAIR=FLuid Attenuation Inversion Recovery, MS=Multiple Sclerosis.}
    }
    \label{fig:LES_CHANGE}
\end{figure*}

Fig.~\ref{fig:APCs_MS} shows the same experiment
for the 100 stable vs.~the 100 progressive MS patients of the Munich dataset.
Both SAMSEG-Long and SAMSEG yielded large 
differences
in the atrophy trajectories of the cerebellum cortex, amygdala, hippocampus and thalamus, 
with SAMSEG-Long yielding 
nominally
higher effect sizes \revision{and 
smaller sample sizes (higher power)%
} compared to SAMSEG in these structures 
(\revision{Cohen's d: }[0.37-0.53 vs.~0.33-0.50]\revision{, 
sample sizes:
[88-184 vs.~101-226]}).
These results are in line with previous studies showing more marked atrophy trajectories in progressive MS patients than stable MS patients~\citep{Eshaghi2018, Cagol2022}.
%
We also report in Fig.~\ref{fig:LES_CHANGE} the yearly volume increase and decrease of lesions 
\revision{%
-- 
as defined in Sec.~\ref{sec:metricsAndMeasures}, i.e., ignoring simultaneous shrinkage and growth, respectively
--
in stable and progressive MS patients, both for SAMSEG-Long and LST-Long. 
}%
(Note that we cannot report such results for SAMSEG, as its segmentations are not longitudinally registered, therefore not allowing voxelwise lesion comparisons.) 
\revision{%
Similar to the findings in~\citep{Pongratz2019}, 
where lesion volume increase and decrease were found to be comparable in size but larger in more active patients,
%
both }
SAMSEG-Long and LST-Long 
detected more lesion changes \revision{(in both directions, i.e., lesion growth and shrinkage)}
in the progressive patients compared to the stable ones.
%
%
\revision{Both methods yielded} 
similar effect sizes between the two patient groups%
\revision{, but}
in absolute terms the lesion volume changes estimated by LST-Long were an order of magnitude smaller than the ones computed by SAMSEG-Long (0.07-0.25 ml/year vs.~1.4-1.7 ml/year).
Considering also the almost perfect test-retest performance of LST-Long in Sec.~\ref{subsec:TestRetest}, it seems that the method may be over-regularizing over time.

\begin{figure*}[!ht]
    \centering
    \includegraphics[width=0.49\linewidth]{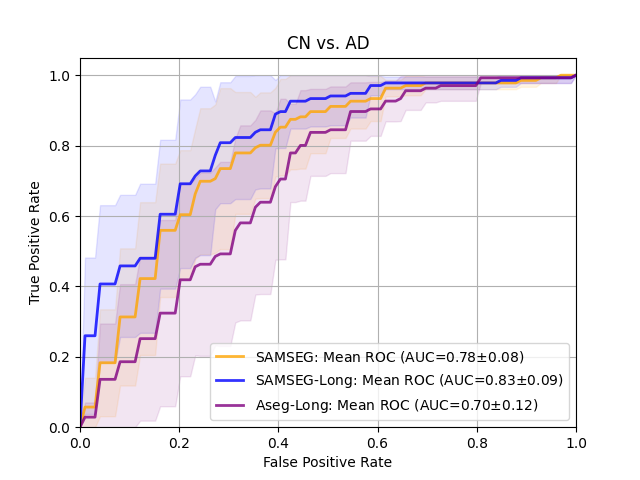}
    \includegraphics[width=0.49\linewidth]{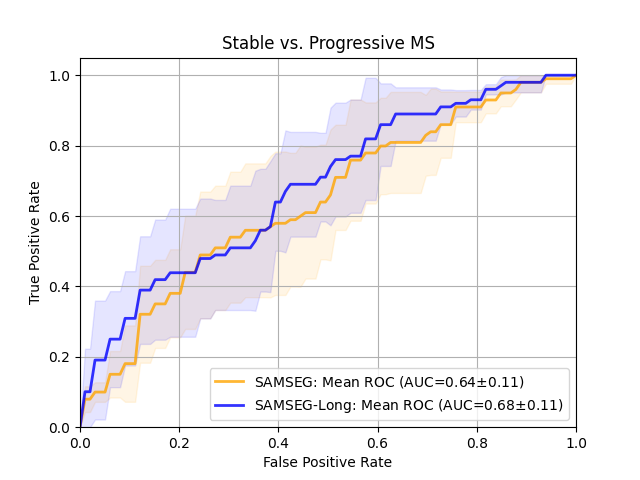}
    \caption{
    ROC curves for an LDA classifier trained on APC values computed from the T1w scans of the ADNI (CN=66, AD=64) and OASIS (CN=72, AD=64) datasets (left) and from the T1w and FLAIR scans of the Munich dataset (MS stable=100, MS progressive=100) (right).
    LDA classifiers were trained and tested using a 5-fold cross-validation procedure. 
    For each method, the area under the curve (AUC) is reported, the mean ROC curve is displayed as a solid line, and the shaded area represents $\pm1$ standard deviation error from the mean ROC curve.
    \revision{ROC=receiver operating characteristic curves, LDA=linear discriminant analysis, APC=Annualized Percentage Change, CN=Cognitive Normal, AD=Alzheimer's Disease, MS=Multiple Sclerosis, T1w=T1-weighted, FLAIR=FLuid Attenuation Inversion Recovery, MS=Multiple Sclerosis.}
    }

    \label{fig:ROC_analysis}
\end{figure*}

As a final experiment to compare the various methods' ability to detect longitudinal disease effects, we assessed whether their APC values contain enough information to correctly classify individual subjects into their respective population groups. In contrast to previous experiments that focused on each structure independently, here we utilized all structures simultaneously:
For each method, we trained and tested a linear discriminant analysis (LDA) classifier on APC values using a 5-fold cross-validation procedure. For each fold, 80\% of the data was used for training and the remaining 20\% for testing.
Fig.~\ref{fig:ROC_analysis} (left) shows receiver operating characteristic curves (ROC) obtained by training LDA classifiers on APC values computed from the ADNI and OASIS dataset for CN and AD patients.
The LDA classifier trained on APC values computed by SAMSEG-Long achieved the highest area under the curve (0.83), followed by the classifiers trained on SAMSEG and Aseg-Long APC values (SAMSEG: 0.78, Aseg-Long: 0.70).
Using the same cross-validation procedure, we also trained an LDA classifier on the APC values computed from the Munich dataset for stable and progressive MS patients, and reported ROC curves in Fig.~\ref{fig:ROC_analysis} (right). 
In line with the previous experiment, the LDA classifier trained on SAMSEG-Long APC values obtained a higher area under the curve compared to the classifier trained on SAMSEG APC values (0.68 vs.~0.64).  

\revision{
\subsection{Longitudinal lesion segmentation performance}

In order to 
directly
compare automatic longitudinal lesion segmentations 
against ``ground truth'' lesion annotations performed by human experts, 
we analyzed the 14 MS patients with longitudinal scans provided by the ISBI dataset. 
Since the heavy preprocessing applied to this data proved problematic for LST-Long, we were unable to get results with this method, 
and therefore only report performance for SAMSEG and SAMSEG-Long.

The ISBI challenge website\footnote{\url{https://smart-stats-tools.org/lesion-challenge}} 
allows users to upload 
lesion segmentation masks,
and ranks 
submissions according to an overall lesion segmentation performance score that takes into account Dice overlap, volume correlation, surface distance, and a few other metrics
against manual annotations that 
remain hidden 
(see~\citep{Carass2017} for details). A score of 100 indicates perfect correspondence, while 90 is meant to correspond to human inter-rater performance~\citep{Carass2017, Styner2008}. SAMSEG obtained a score of 88.31, while SAMSEG-Long 
similarly scored
88.61. 
The 
Dice 
coefficients
between 
the
manual and 
the 
corresponding automatic lesion 
segmentations 
-- 
computed for each rater, 
subject and 
time point individually --  
are also provided by the website, and
are summarized in Fig.~\ref{fig:diceISBI}.
The median Dice score was around 0.58 for both SAMSEG and SAMSEG-Long, and 
no statistical significance was found between the two methods.

When interpreting these results, it is worth pointing out that,
although the 
ISBI 
data
itself 
is longitudinal, the \emph{manual annotations} are not: 
As detailed in~\citep{Carass2017},
the human raters were presented with 
the scan of each time point
independently, 
resulting in poor longitudinal consistency
in their manual annotations.
Furthermore,
both SAMSEG and SAMSEG-Long were used ``out of the box'' 
without any form of additional tuning.
This
should be taken into account when comparing 
their 
numerical scores
against those obtained by methods 
that were 
specifically 
%
optimized
for 
the 
exact
scanner and imaging protocol 
of the challenge
(typically by training on the matched training data that is also provided by the challenge).

\begin{figure}
    \centering
    \includegraphics[width=0.4\textwidth]{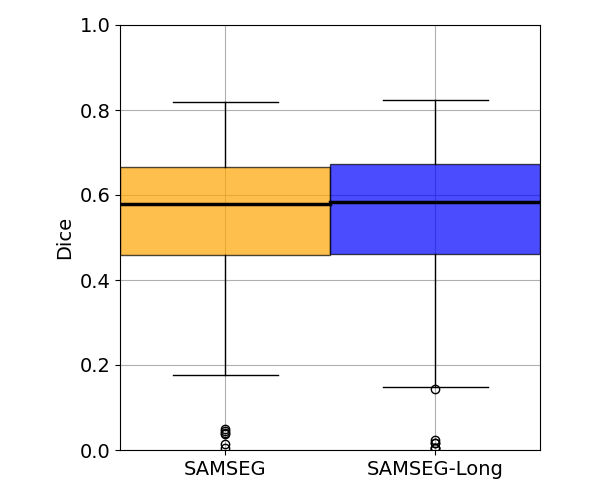}
    \caption{\revision{Longitudinal lesion segmentation performance in terms of Dice overlaps with manual raters computed from T1w and FLAIR scans on the 14 MS subjects of the ISBI dataset for SAMSEG and SAMSEG-Long. 
    Each automatic segmentation is compared against each of the two manual segmentations provided in the dataset.
    No statistically significant difference was detected between the two methods using a two-tailed paired t-test.
    T1w=T1-weighted, FLAIR=FLuid Attenuation Inversion Recovery, MS=Multiple Sclerosis.}
    }
    \label{fig:diceISBI}
\end{figure}

}


\section{Discussion and conclusion}
\label{sec:Discussion}

In this paper we have proposed and evaluated a new method for
segmenting dozens of neuroanatomical structures from longitudinal MRI scans.
Temporal regularization is achieved by introducing a set of subject-specific latent variables in an existing cross-sectional segmentation method. An extension for segmenting white matter lesions is also available,
allowing users to simultaneously track lesion evolution and morphological changes in various brain structures in e.g., patients suffering from MS.
The proposed method 
does not make any 
assumptions on the scanner, the MRI protocol, or the number and timing of longitudinal follow-up scans,
and
is publicly available as part of the open-source neuroimaging package FreeSurfer.

Our
experiments 
indicate that the proposed method has better test-retest reliability compared to benchmark methods, and that it is more sensitive to disease-related changes in AD and MS. In other words, our new tool generated results that were both more sensitive and more specific -- suggesting that its use may 
bring
several advantages, such as the need for fewer subjects in longitudinal studies, a better stratification of patients, and more precise evaluation of treatment efficacy.

%
%

\revisionTwo{
%
%
The robustness and generalizability of the 
method 
across different scanner platforms, field strengths, acquisition protocols and image resolutions
was demonstrated by its successful ``out-of-the-box''
application 
(i.e., without any  form of retraining or retuning)
on a diverse set of longitudinal datasets.
These datasets 
included single- and multi-contrast longitudinal scans
with a range of time gaps and total number of time points, 
from both healthy and diseased subjects,
comprising over 4,500  MRIs in total.
%
The cross-sectional methods the proposed technique builds upon have 
themselves 
previously
been 
validated, 
by 
comparing their
segmentations
against those of 
manual raters
on images acquired with different scanners and imaging protocols~\citep{Puonti2016,Cerri2021}.
%
Although this provides evidence of 
%
the generalizability 
of the method,
a direct evaluation of its 
actual
\emph{longitudinal}
segmentation performance
has been hampered by
the lack of 
manual
longitudinal annotations 
that could serve as ``ground truth''.
%
%
%
As is 
common in the 
longitudinal segmentation
literature,
we therefore resorted to an \emph{indirect}
validation and 
tested the ability 
of the 
method to detect disease-related temporal changes
instead.
This, however, 
has
the 
limitation
that
higher sensitivity to group differences does not necessarily imply anatomically more correct segmentations. 
}

Although 
we believe 
the proposed tool
will be 
helpful
for
researchers and clinicians investigating temporal-morphological changes in the brain,
the method still has several limitations.
First, 
the method currently only produces volumetric whole-brain segmentations,
as opposed to the longitudinal tool of~\citep{Reuter2012} that additionally also computes detailed longitudinal parcellations of the cortical surface.
%
%
\revisionTwo{%
Second, although our tool can segment white matter lesions from longitudinal scans
acquired with conventional MRI sequences, 
the signal changes that it detects in such images are 
nonspecific,
with 
several
different 
processes 
all
resulting in similar MRI intensity profiles~\citep{Sethi2017,Pongratz2019}.
Disentangling the 
various
underlying 
pathological changes due to e.g., demyelination, remyelination, inflammation or edema may 
ultimately
become 
feasible
with more advanced MRI techniques~\citep{Pirko2008, Oh2019}, 
but will 
likely
require 
further development
of the image analysis techniques described here.%
%
%
%
%
%
%
%
%
}
%
%
%
%
%
%
\revisionTwo{Third,}
the method does not currently exploit the time dimension explicitly to constrain~\citep{Metcalf1992, Welti2001, Solomon2004} or analyze lesion evolution~\citep{Thirion1999, Gerig2000, Rey2002, Bosc2003, Elliott2013} in detail. 
Dedicated methods for detecting \emph{new} or \emph{growing} lesions by comparing two consecutive time points, in particular, have received ongoing attention\revision{~\citep{Commowick2021, Diaz2022}: Many methods rely on image subtraction techniques~\citep{Sweeney2013, Battaglini2014, Ganiler2014, McKinley2020, Sepahvand2020, Kruger2020, Klistorner2021}, while others use spatial deformation information~\citep{Elliott2019, Preziosa2022}}. 
Although 
\revision{%
this type of 
functionality
is
not directly 
provided
by}
the proposed tool,
its ability to tightly standardize longitudinal images -- both in terms of removing global intensity scaling differences and bias field artifacts, and in terms of establishing accurate longitudinal nonlinear registrations across the various time points --
may be leveraged 
to further develop such methods.

\section{Acknowledgments}
\label{sec:Acknowledgments}

This project has received funding from the European Union's Horizon 2020 research and innovation program under the Marie Sklodowska-Curie grant agreement No 765148,
as well as from the National Institute Of Neurological Disorders and Stroke under project number R01NS112161.
Douglas N. Greve has received funding from the National Institute Of Neurological Disorders and Stroke (R01NS105820), the National Institute for Biomedical Imaging and Bioengineering (R01EB023281), and the National Institute on Aging (R01AG057672, R01AG059011, U19AG068753).
Henrik Lundell has received funding from the European Research Council (ERC) under the European Union’s Horizon 2020 research and innovation programme (grant agreement No 804746).
Hartwig R. Siebner holds a 5-year professorship in precision medicine at the Faculty of Health Sciences and Medicine, University of Copenhagen which is sponsored by the Lundbeck Foundation (Grant Nr. R186-2015-2138). 
Mark M\"{u}hlau was also supported by the German Research Foundation (Priority Program SPP2177, Radiomics: Next Generation of Biomedical Imaging) -- project number 428223038.
Images of MS patients from the Munich dataset were subgroups of the MS cohort study of the Technical University of Munich (TUM-MS) run by the Dept. of Neurology in close collaboration with the Dept. of Diagnostic and Interventional Neuroradiology. A particular thanks to Claus Zimmer, Bernhard, Hemmer, Jan Kirschke, Achim Berthele, Benedikt Wiestler, and Matthias Bussas for their ongoing support.

Data used in preparation of this article were obtained from the Alzheimer's Disease Neuroimaging Initiative (ADNI) database (\url{http://adni.loni.usc.edu}). As such, the investigators within the ADNI contributed to the design and implementation of ADNI and/or provided data but did not participate in analysis or writing of this report. A complete listing of ADNI investigators can be found at \url{adni.loni.usc.edu/wp-content/uploads/how_to_apply/ADNI_Acknowledgement_List.pdf}.

\section{Conflicts of interest}
\label{sec:conflictsOfInterest}

Hartwig R. Siebner has received honoraria as speaker from Lundbeck AS, Denmark, Sanofi Genzyme, Denmark and Novartis, Denmark, as consultant from Sanofi Genzyme, Denmark and Lundbeck AS, Denmark, and as editor-in-chief (Neuroimage Clinical) and senior editor (NeuroImage) from Elsevier Publishers, Amsterdam, The Netherlands. He has received royalties as book editor from Springer Publishers, Stuttgart, Germany and from Gyldendal Publishers, Copenhagen, Denmark.

\bibliography{main.bib} 

\appendix
\section{Datasets details}
\label{app:datasets}
  
We use disparate datasets for validating the proposed longitudinal method against benchmark methods, as well as for tuning the hyperparameters of the model. We here describe them in detail:
\begin{itemize}

    \item \textbf{MIRIAD-TR-HT}~\citep{Malone2013}: This dataset consists of test-retest scans of 10 healthy elderly people and 30 AD patients from the Minimal Interval Resonance Imaging in Alzheimer's Disease (MIRIAD) project\footnote{\url{https://www.nitrc.org/projects/miriad/}}.
    T1-weighted (T1w) scans with voxel size of $0.9375\times0.9375\times1.5$ mm were acquired on a GE Signa 1.5T scanner using an Inversion Recovery prepared - Fast SPoiled Gradient Recalled (IR-FSPGR) sequence. For each subject, 2 test-retest images were acquired without removing the patient from the scanner, for a total of 80 scans. 

    \item \textbf{ADNI-HT}: This dataset consists of T1w longitudinal scans of 80 subjects from the ADNI project\footnote{\url{http://adni.loni.usc.edu/}}. 
    Different 3T scanners from multiple sites were used for scanning subjects.
    For each subject, T1w scans were acquired using an IR-FSPGR or MP-RAGE sequence at different image resolution ([min-max]: [1.18-1.21] $\times$ [0.92-1.31] $\times$ [0.93-1.29] mm), with minor resolution variations between follow up scans ($\leq 0.01$ mm in each axis).
    Subjects were scanned on average 3.56 times (minimum: 2 times; maximum: 5 times; 285 scans in total), with a mean interval between scans equal to 313 days (minimum: 107 days; maximum: 1121 days).
    The mean age at baseline was 75 years (minimum: 57; maximum: 90).
    Subjects were divided into two groups: cognitive normal (CN) (n=37) and AD (n=53). 

    \item \textbf{MIRIAD-TR}~\citep{Malone2013}: This dataset consists of test-retest scans of 13 healthy elderly people and 16 AD patients from the MIRIAD project -- distinct from the subjects of the MIRIAD-TR-HT dataset. For each subject, 2 test-retest images were acquired at 2 or 3 different times without removing the patient from the scanner, for a total of 146 scans.

    \item \textbf{ADNI}: This dataset consists of T1w longitudinal scans of 130
    subjects from the ADNI project -- distinct from the subjects of the ADNI-HT dataset. 
    Different scanners from multiple sites and multiple field strengths (1.5T and 3T) were used for scanning subjects.
    For each subject, T1w scans were acquired using an IR-FSPGR or MP-RAGE sequence
    at different image resolutions ([min-max]: [1.17-1.21] $\times$ [0.91-1.31] $\times$ [0.92-1.31] mm), with minor resolution variations between follow-up scans ($\leq 0.01$ mm in each axis).
    Subjects were scanned on average 3.70 times (minimum: 2 times; maximum: 5 times; 477 scans in total), with a mean interval between scans equal to 298 days (minimum: 65 days; maximum: 903 days). The mean age at baseline was 76 years (minimum: 57; maximum: 89).
    Subjects were divided into two groups: CN (n=66) and AD (n=64). 

    \item \textbf{OASIS}~\citep{Marcus2010}: This dataset consists of T1w longitudinal scans of 136 subjects from the Open Access Series of Imaging Studies (OASIS-2) project\footnote{\url{https://www.oasis-brains.org/}}. 
    T1w scans with voxel size of $1\times1\times1.25$ mm were acquired on a 1.5T Siemens Vision scanner using a MP-RAGE sequence. 
    Subjects were scanned on average 2.47 times (minimum: 2 times; maximum: 5 times; 336 scans in total), with a mean interval between scans equal to 702 days (minimum: 182 days; maximum: 1510
    days). 
    The mean age at baseline was 75 years (minimum: 60; maximum: 96).
    Scans were acquired on a Siemens Vision 1.5T scanner. Subjects were divided into 
    two groups: CN (n=72) and AD (n=64). 
    
    \item \textbf{OASIS-TR}~\citep{Marcus2010}: This dataset consists of the same subjects of the OASIS dataset plus 14 additional subjects from the same project that were diagnosed as converted, for a total of 150 subjects. For each subject, 3 or 4 individual T1w MRI scans obtained in single scan sessions were acquired, for a total of 1845 scans.

    \item \textbf{Munich-TR}~\citep{Biberacher2016}: This dataset consists of longitudinal T1w and FLuid Attenuation Inversion Recovery (FLAIR) scans of 2 MS subjects. For each subject, 6 repeated scans were acquired from 3 different 3T scanners (Philips Achieva; Siemens Verio; GE Signa MR750) within 3 weeks (mean interval between successive scans equal to 3 days (minimum: 2 days; maximum: 7 days).
    Voxel sizes for the T1w scans were $1\times1\times1$ mm for the GE scanner, $1\times1\times1$ mm for the Philips scanner and $1.1\times1.1\times1$ mm for the Siemens scanner. Voxel sizes for the FLAIR scans were $1\times1\times1$ mm for the GE scanner, $1\times1\times1.5$ mm for the Philips scanner, and $1\times1\times1$ mm for the Siemens scanner.
    T1w scans were acquired using a IR-FSPGR or MP-RAGE sequence.
    Two scans of ``Subject-1'' were excluded from the dataset due to scanning protocol violations~\citep{Biberacher2016}, and they are not included in the experiments. 
    
    \item \textbf{Munich}: This dataset consists of longitudinal T1w (MP-RAGE sequence) and FLAIR scans of 200 MS subjects acquired on a Philips Achieva 3T scanner.
    Subjects were scanned on average 6.45 times (minimum: 2 times; maximum: 24 times; 1289 scans in total), with a mean interval between scans equal to 353 days (minimum: 18 days; maximum: 3287 days) and at least 365 days between the first and last visit.
    Voxel sizes for the T1w scans were either $0.75\times0.75\times0.75$ or, after a scanner change, $1\times1\times1$ mm, while voxel sizes for the FLAIR scans were either $0.75\times0.75\times0.75$ or $0.9\times0.9\times1.5$ mm, respectively.
    Because some of the T1w scans have image resolutions that differ between time points of the same subject, we resampled all the T1w scans of this dataset to 1mm isotropic resolution before using them in the experiments.
    The mean age at baseline was 40 years (minimum: 18; maximum: 74).
    At each subject's visit, an Expanded Disability Status Scale (EDSS) score was assessed (mean: 2.46; minimum: 0; maximum: 8.5).
    Subjects were divided into two groups: stable (n=100) and progressive (n=100) according to the following three strata criteria~\citep{Lorscheider2016}: Patients were classified as progressive if 
    (1) the baseline EDSS score is 0 and there is an increase of 1.5 points from the baseline EDSS score to the last time point EDSS score;
    (2) the baseline EDSS score is between 1 and 5.5 and there is an increase of 1 point from the baseline EDSS score to the last time point EDSS score;
    (3) the baseline EDSS score is above 5.5 and there is an increase of 0.5 points from the baseline EDSS score to the last time point EDSS score.
    Patients were classified as stable otherwise.

    \revision{\item \textbf{ISBI}: This dataset is the publicly available test set of the MS lesion segmentation challenge that was held at the 2015 International Symposium on Biomedical Imaging~\cite{Carass2017}. It consists of 14 longitudinal MS cases, scanned on average 4.36 times (minimum: 4 times, maximum: 6 times), separated by approximately one year (mean: 391 days, minimum: 299 days, maximum: 503 days). 
    All images were acquired on the same Philips 3T scanner, using a 3D T1w, T2w, PDw and FLAIR sequence.
    Voxel sizes were $0.82\times0.82\times1.17$ mm for the T1w scans, and 
    $0.82\times0.82\times2.22$ mm
    for T2w, PDw, and FLAIR.
    Images were first preprocessed (inhomogeneity correction, skull stripping, dura stripping, and again inhomogeneity correction -- see~\citep{Carass2017} for details). Each T1w baseline image was then registered to a 1 mm MNI template and used as target image for registering successive time point images.
    Each subject's lesions were delineated by two different raters on the FLAIR scan, and, if necessary, corrected using the other contrasts. Note that the raters were presented with each scan independently, 
    without aiming for longitudinal consistency in their segmentations.
    In our experiments we used the input combination T1w-FLAIR, which we have found to lead to 
    good
    lesion segmentation performance in a previous study~\cite{Cerri2021}.} 
    
\end{itemize}

\section{\revision{Additional results}}

\begin{figure*}[!ht]
    \centering
    \begin{minipage}[t]{.99\linewidth}
        \includegraphics[width=\textwidth]{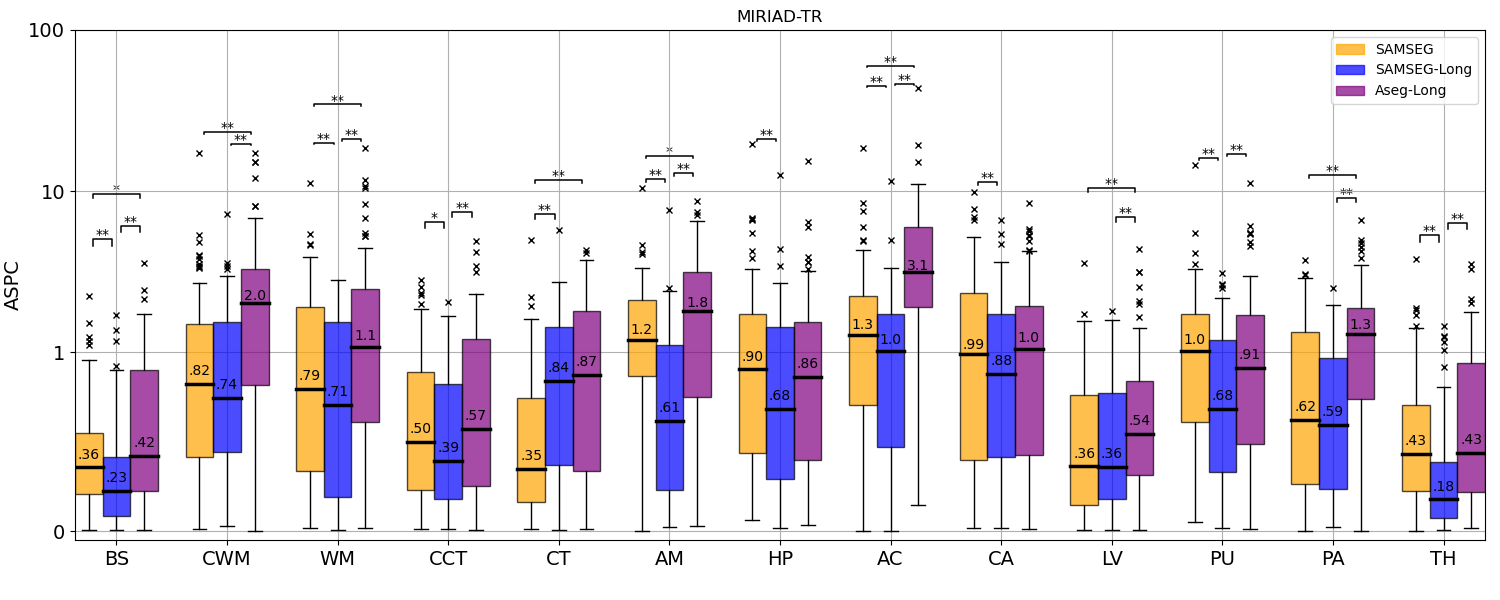}
    \end{minipage}
    \begin{minipage}[t]{.99\linewidth}
        \includegraphics[width=\textwidth]{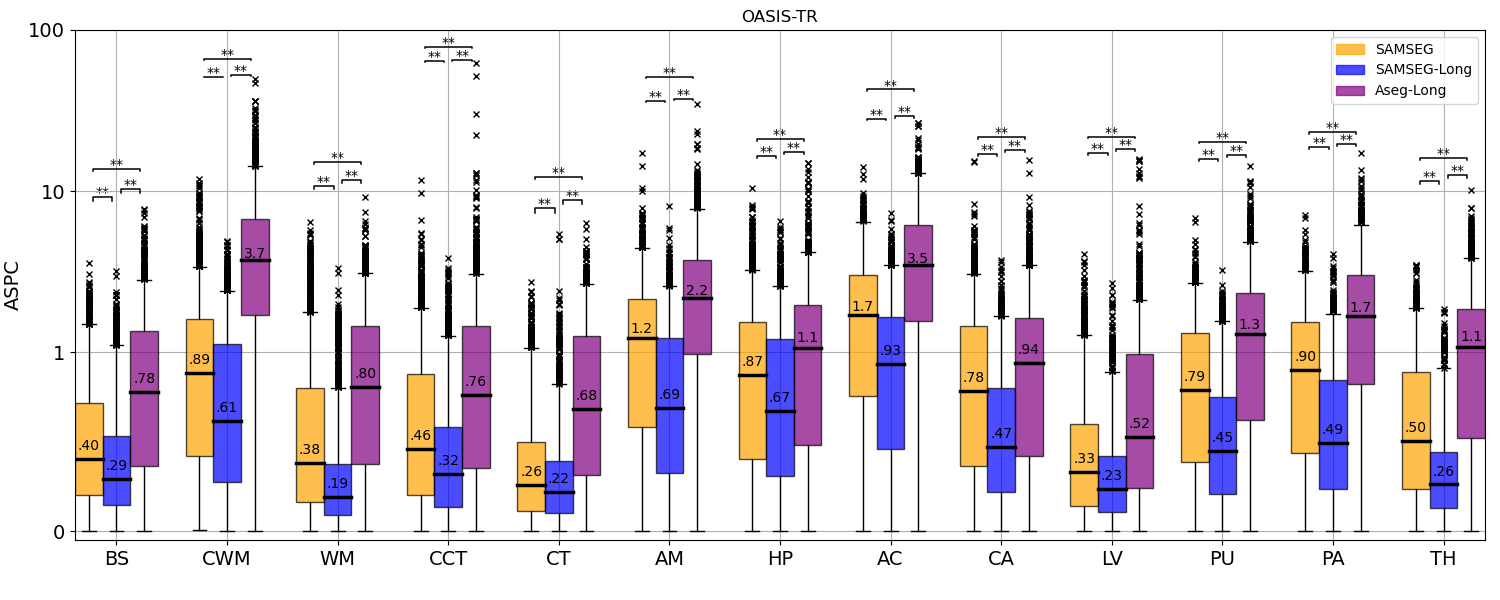}
    \end{minipage}
    \caption{
    ASPCs for the test-retest T1w scans of the MIRIAD-TR (\# scans: 146, AD=13, CN=16) and OASIS-TR (\# scans: 1845, AD=72, Converted=14, CN=64) datasets for different brain structures for SAMSEG, SAMSEG-Long and Aseg-Long.
    Within each boxplot, the median \revision{ASPC} value is also reported.
    Statistically significant differences between two methods, were computed with a Wilcoxon signed-rank test, and indicated by asterisks 
    \revision{(``**'' for p-value $<$ 0.01 and ``*'' for p-value $<$ 0.05).}
    \revision{ASPC=Absolute Symmetrized Percent Change, T1w=T1-weighted, CN=Cognitive Normal, AD=Alzheimer's Disease,}
    \revision{BS=brain stem, CWM=cerebellum white matter, WM=cerebral white matter, CCT=cerebellum cortex, CT=cerebral cortex, AM=amygdala, HP=hippocampus, AC=nucleus accumbens, CA=caudate, LV=lateral ventricle, PU=putamen, PA=pallidum, TH=thalamus.}
    }
    \label{fig:ASPCsMIRIAD_OASIS}
\end{figure*}

\begin{figure*}[!ht]
    \begin{center}
    \centering
    \begin{minipage}[t]{.99\linewidth}
      \includegraphics[height=6.5cm]{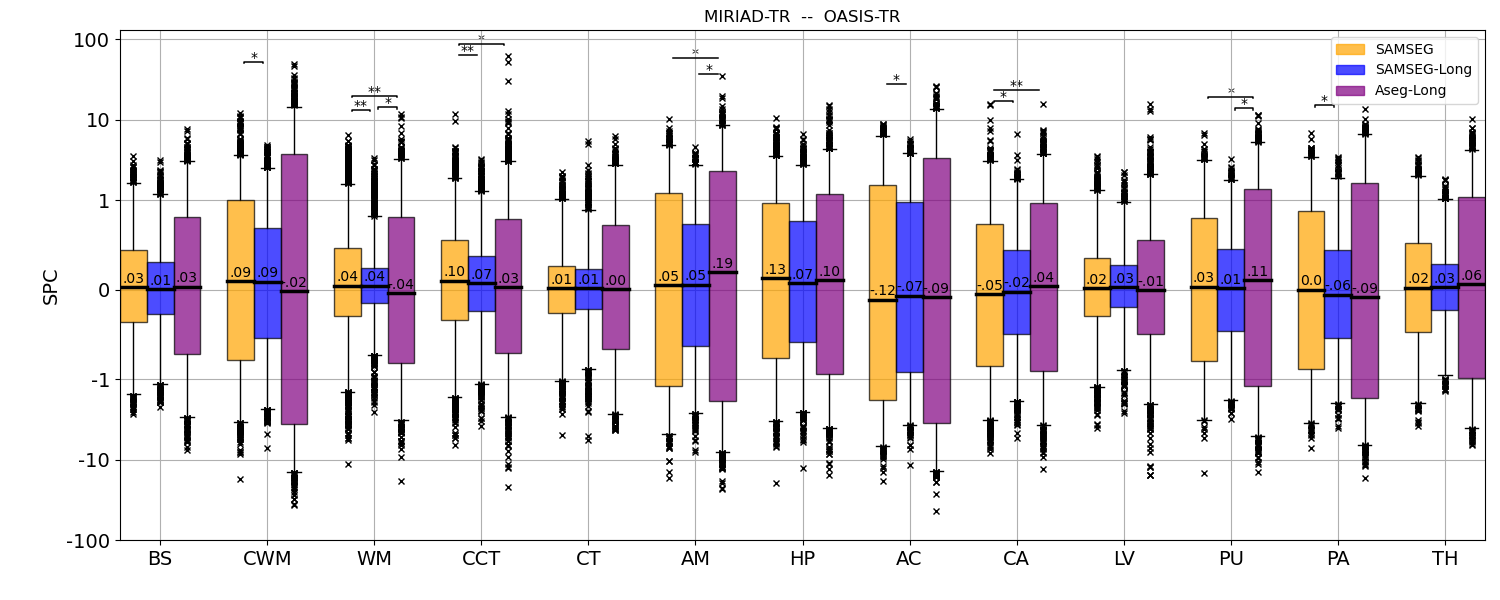}
    \end{minipage}
    \begin{minipage}[t]{.99\linewidth}
        \includegraphics[height=6.5cm]{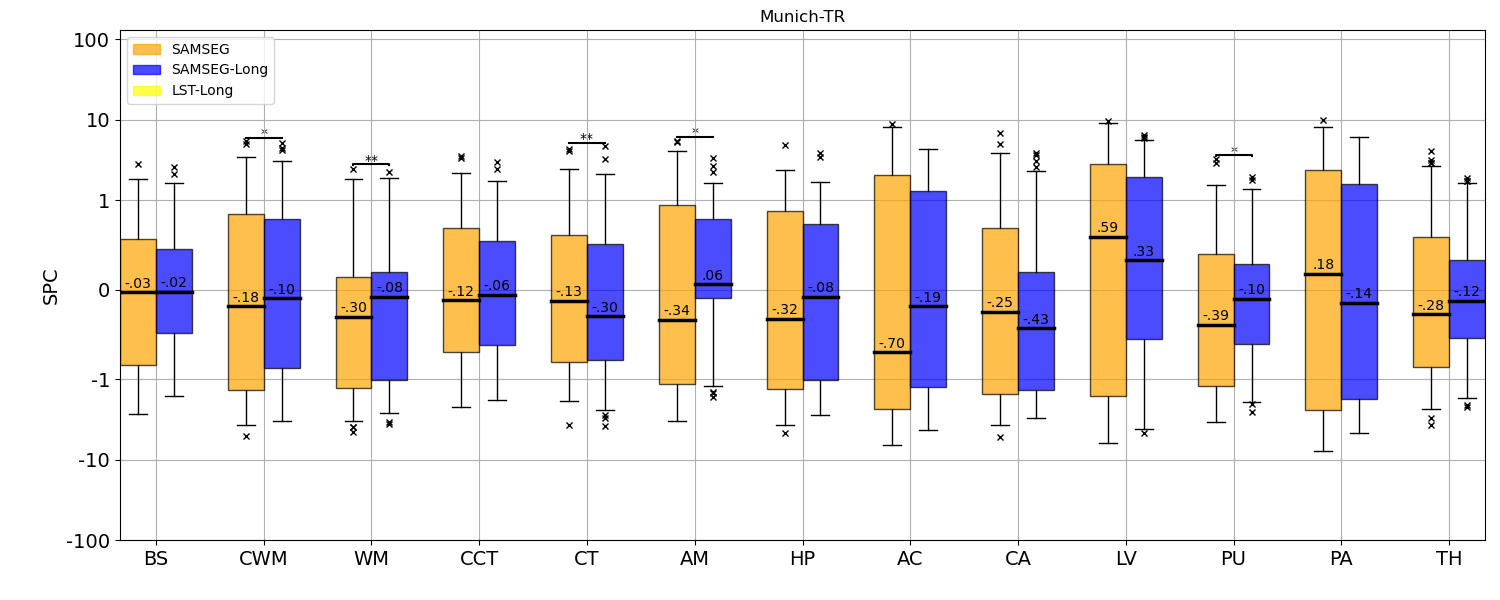}
        \hspace{-0.35cm}
        \includegraphics[height=6.5cm]{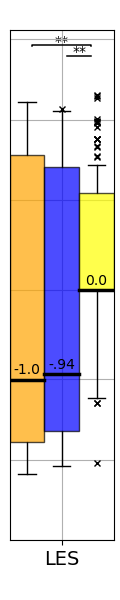}
    \end{minipage}    
    \end{center}
    \caption{
    \revision{%
    SPC values for the test-retest T1w scans of the combined MIRIAD-TR (\# scans: 146, AD=13, CN=16) and OASIS-TR (\# scans: 1845, AD=72, Converted=14, CN=64) datasets (top) and for the test-retest T1w-FLAIR scans of the Munich-TR (\# scans: 34, MS=2) dataset (bottom). SPCs values were computed for different brain structures and white matter lesion (LES) for the proposed method and the benchmark methods.
    Within each boxplot, the median SPC value is also reported. 
    Statistically significant differences between two methods were computed with a Wilcoxon signed-rank test, and are indicated by asterisks (``**'' for p-value $<$ 0.01, and ``*'' for p-value $<$ 0.05). 
    SPC=Symmetrized Percent Change, T1w=T1-weighted, FLAIR=FLuid Attenuation Inversion Recovery, AD=Alzheimer's Disease, CN=Cognitive Normal, MS=Multiple Sclerosis,
    BS=brain stem, CWM=cerebellum white matter, WM=cerebral white matter, CCT=cerebellum cortex, CT=cerebral cortex, AM=amygdala, HP=hippocampus, AC=nucleus accumbens, CA=caudate, LV=lateral ventricle, PU=putamen, PA=pallidum, TH=thalamus, LES=white matter lesion.
    }
    }
    \label{fig:SPCs}
\end{figure*}

\begin{figure*}[!ht]
    \centering
    \begin{minipage}[t]{.95\linewidth}
        \includegraphics[width=\textwidth]{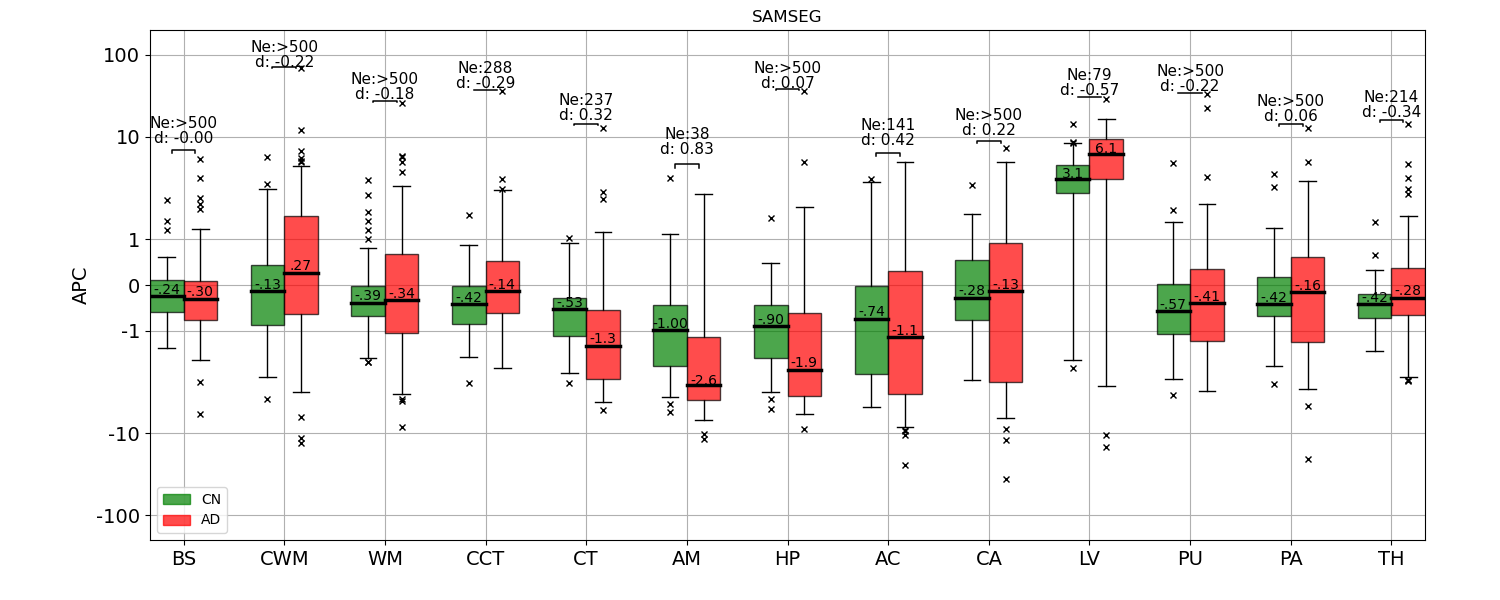}
    \end{minipage}
    \begin{minipage}[t]{.95\linewidth}
        \includegraphics[width=\textwidth]{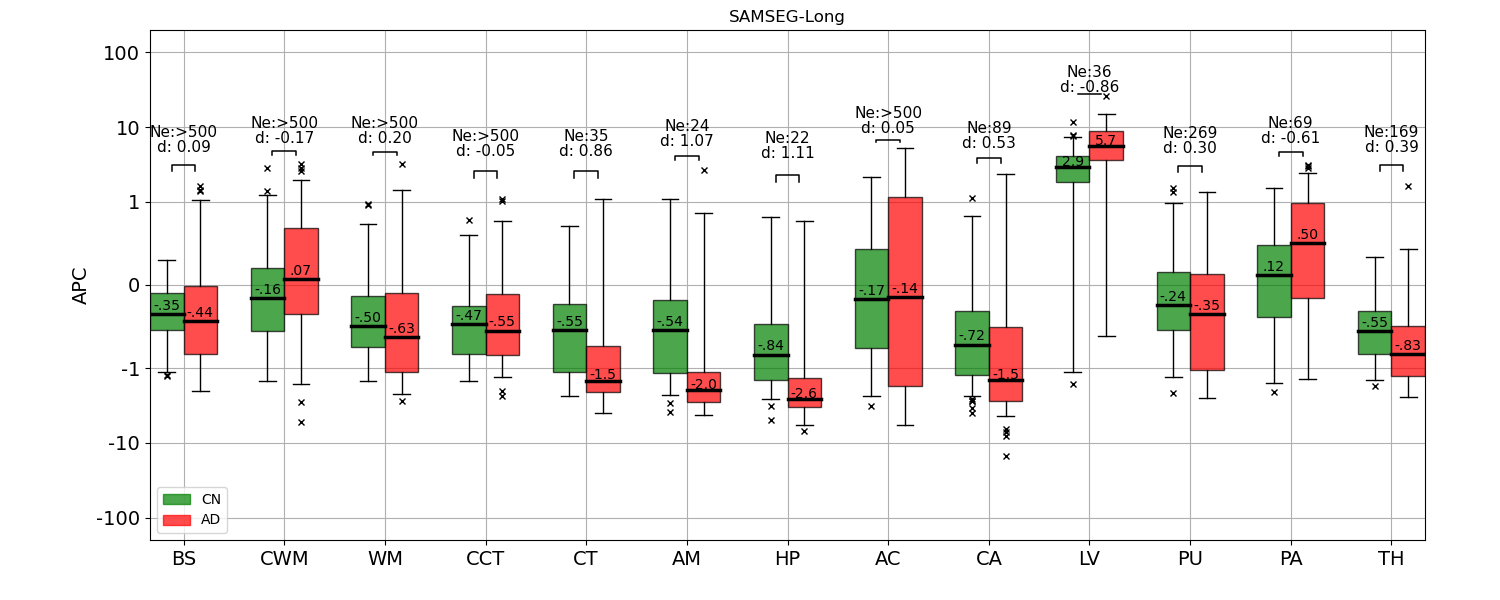}
    \end{minipage}
        \begin{minipage}[t]{.95\linewidth}
        \includegraphics[width=\textwidth]{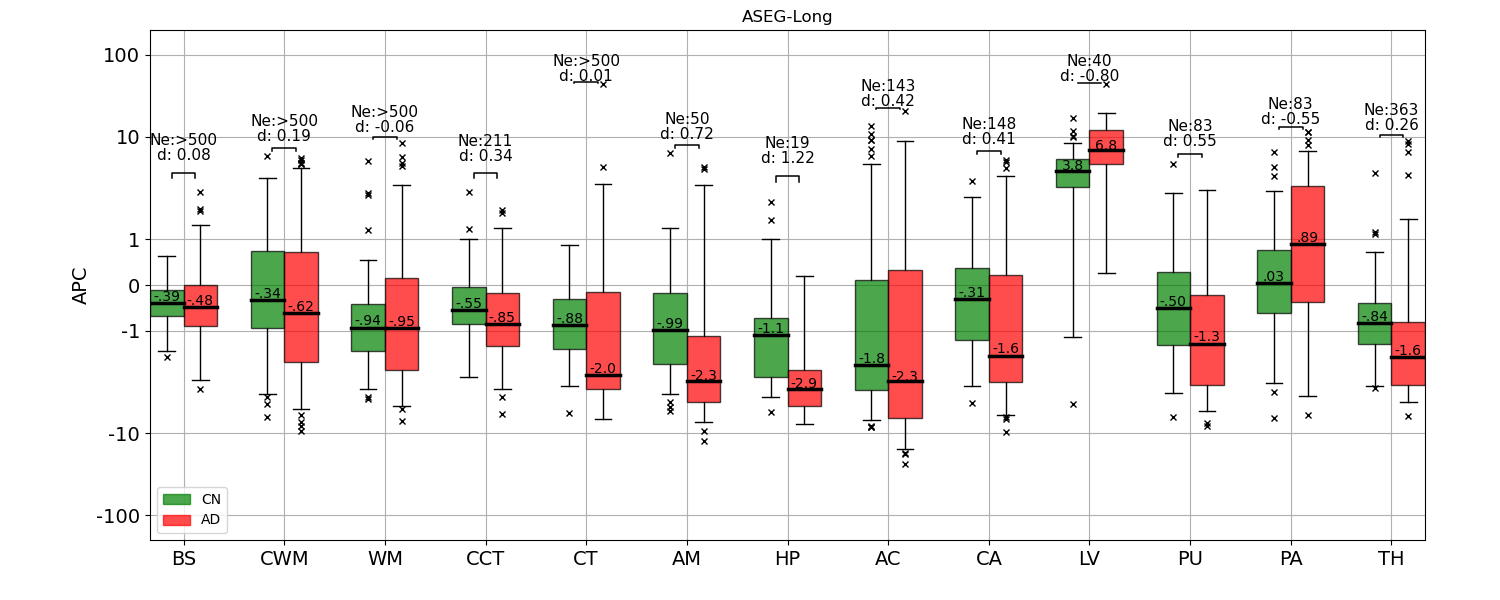}
    \end{minipage}
    \caption{
    APCs computed from the T1W scans of the 130 subjects of the ADNI dataset (CN=66, AD=64) for SAMSEG, SAMSEG-Long, and Aseg-Long.
    Cohen's d effect size\revision{ (d) and effective number of subjects (Ne) computed from a power analysis (80\% power, 0.05 significance level) are} reported above each pair of box plots.
    Within each boxplot, the median \revision{APC} value is also reported.
    \revision{APC=Annualized Percentage Change, CN=Cognitive Normal, AD=Alzheimer's Disease, T1w=T1-weighted,}
    \revision{BS=brain stem, CWM=cerebellum white matter, WM=cerebral white matter, CCT=cerebellum cortex, CT=cerebral cortex, AM=amygdala, HP=hippocampus, AC=nucleus accumbens, CA=caudate, LV=lateral ventricle, PU=putamen, PA=pallidum, TH=thalamus.}
    }
    \label{fig:APCs_ADNI}
\end{figure*}

\begin{figure*}[!ht]
    \centering
    \begin{minipage}[t]{.95\linewidth}
        \includegraphics[width=\textwidth]{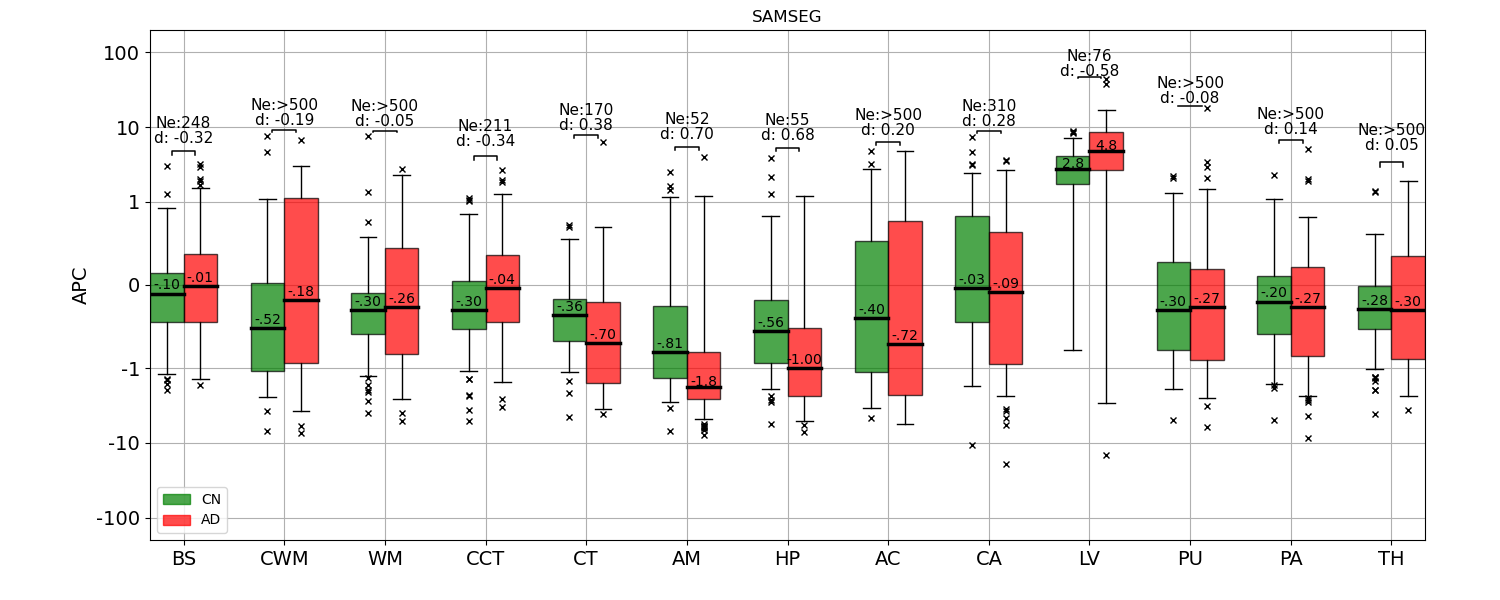}
    \end{minipage}
    \begin{minipage}[t]{.95\linewidth}
        \includegraphics[width=\textwidth]{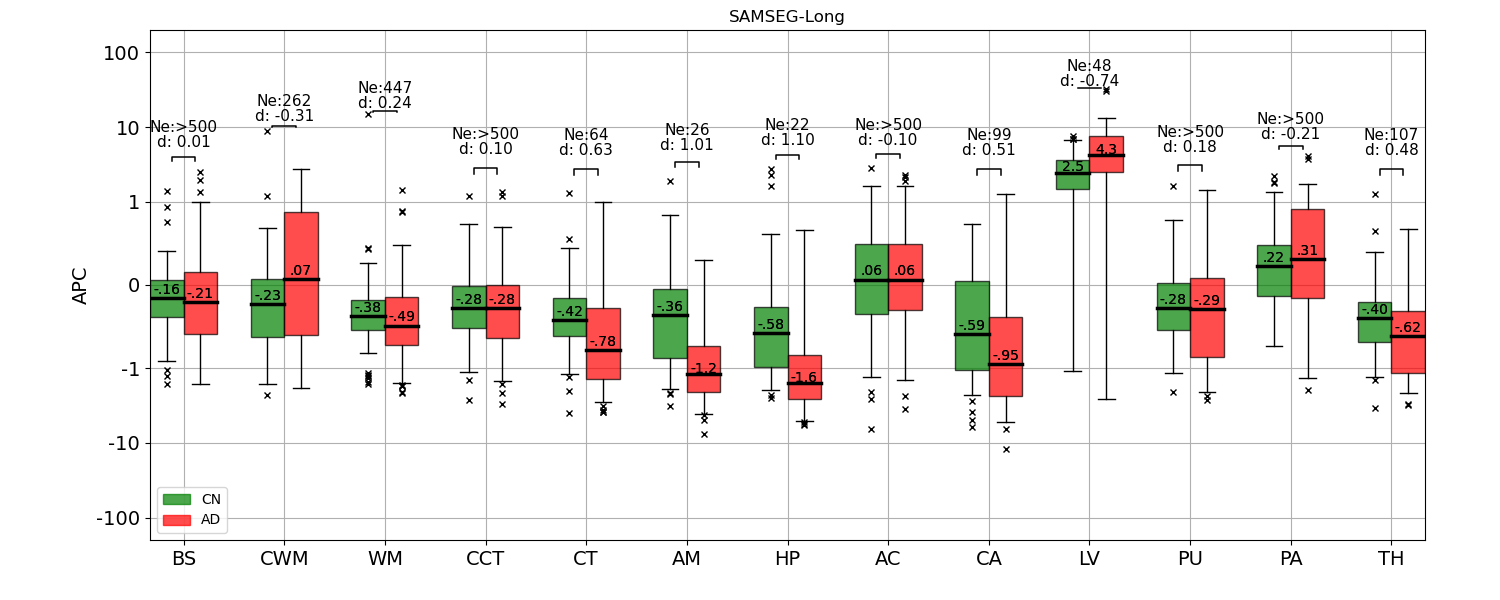}
    \end{minipage}
        \begin{minipage}[t]{.95\linewidth}
        \includegraphics[width=\textwidth]{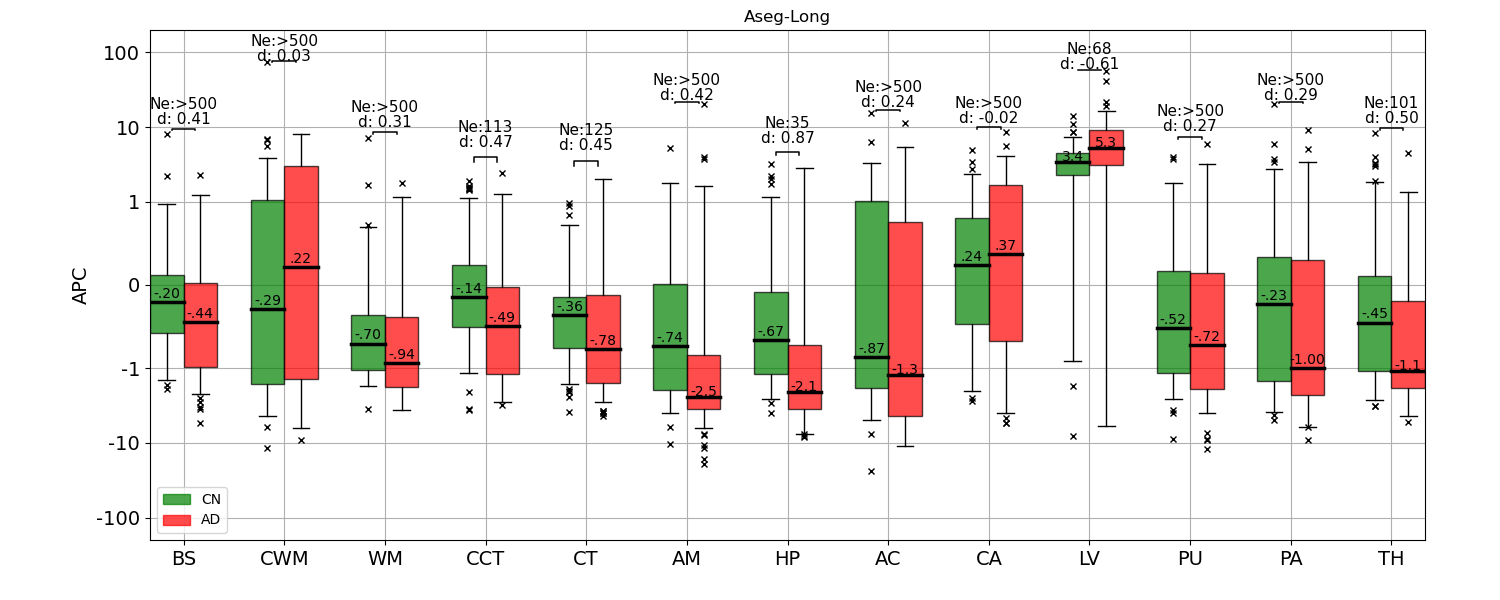}
    \end{minipage}
    \caption{
    APCs computed from the T1W scans of the 136 subjects of the OASIS dataset (CN=72, AD=64) for SAMSEG, SAMSEG-Long, and Aseg-Long.
    Cohen's d effect size\revision{ (d) and effective number of subjects (Ne) computed from a power analysis (80\% power, 0.05 significance level) are} reported above each pair of box plots.
    Within each boxplot, the median \revision{APC} value is also reported.
    \revision{APC=Annualized Percentage Change, CN=Cognitive Normal, AD=Alzheimer's Disease, T1w=T1-weighted,}
    \revision{BS=brain stem, CWM=cerebellum white matter, WM=cerebral white matter, CCT=cerebellum cortex, CT=cerebral cortex, AM=amygdala, HP=hippocampus, AC=nucleus accumbens, CA=caudate, LV=lateral ventricle, PU=putamen, PA=pallidum, TH=thalamus.}
    }
    \label{fig:APCs_OASIS}
\end{figure*}

\end{document}